%% file: main.tex
\documentclass[10pt,twocolumn,letterpaper]{article}

\usepackage{cvpr} 

\input{preamble}

\definecolor{cvprblue}{rgb}{0.21,0.49,0.74}
\usepackage[pagebackref,breaklinks,colorlinks,allcolors=cvprblue]{hyperref}

\def\paperID{*****} 
\def\confName{CVPR}
\def\confYear{2026}

\title{PAI-Studio: Cinematic Video Background Replacement with Camera-Aware Motion}

\author{
  Heyuan Gao\textsuperscript{1,2,*} \quad
  Bangxun Tang\textsuperscript{1,3,*} \quad
  Yiren Song\textsuperscript{1,4,*,\ensuremath{\ddagger}} \quad
  Guian Fang\textsuperscript{1,4} \\
  Zijian He\textsuperscript{1} \quad
  Jie Yang\textsuperscript{1} \quad
  Mike Zheng Shou\textsuperscript{4,\ensuremath{\dagger}} \\
  \\
  \textsuperscript{1}Utopai Studios \quad
  \textsuperscript{2}Nanyang Technological University \\
  \textsuperscript{3}University of California, Irvine \quad
  \textsuperscript{4}Show Lab, National University of Singapore
}
\begin{document}



\twocolumn[{%
\maketitle
\vspace{-1em}
\begin{center}
\animategraphics[width=0.9\textwidth]{5}{teaser_v5_frames/teaser_}{00}{09}
\captionof{figure}{Given a foreground video and background reference images, our method synthesizes a complete video with camera-aware background motion, foreground consistency, and scene-consistent relighting at 1080P resolution. Readers can click and play the video clips in this figure using {\color{red}\textbf{Adobe Acrobat}}.}
\label{fig:teaser}
\end{center}
\vspace{1em}
}]

\begingroup
\renewcommand{\thefootnote}{}
\footnotetext{
\textsuperscript{*}Equal contribution.
\textsuperscript{$\ddagger$}Project leader.
\textsuperscript{$\dagger$}Corresponding author.
}
\endgroup

\begin{abstract}
We present PAI-Studio, a new reference-conditioned video synthesis task that addresses a long-standing challenge in cinematic background replacement: generating dynamic backgrounds aligned with foreground motion while preserving foreground identity, matching reference scene appearance, and achieving globally consistent illumination with realistic foreground relighting. Existing open-source systems and commercial APIs cannot simultaneously ensure motion-consistent background generation, high-fidelity foreground relighting and foreground identity preservation, often resulting in static backgrounds, inconsistent boundaries, and noticeable compositing artifacts. To bridge this gap, we build upon a Diffusion Transformer video backbone and reformulate the problem as an in-context conditional generation task. Through bidirectional attention, our model jointly captures foreground dynamics and background reference information within a unified architecture. We further construct a 30K-scale dataset sourced from high-quality films and online videos to support this task. Extensive evaluations demonstrate that our method significantly outperforms existing open-source and commercial API solutions. Project page: \url{https://github.com/showlab/PAI-Studio}
\end{abstract}


\input{sec/1_intro}

\input{sec/2_related}

\input{sec/3_method}

\input{sec/4_experiment}

\input{sec/5_conclusion}

{
    \small
    \bibliographystyle{ieeenat_fullname}
    \bibliography{main}
}

\input{sec/X_suppl}

\end{document}

%% file: preamble.tex
\usepackage{graphicx}
\usepackage{booktabs}
\usepackage{animate}
\usepackage{caption}
\usepackage{placeins}
\usepackage{float}
\usepackage[table]{xcolor}
\usepackage{colortbl}
\usepackage{multirow}
\usepackage{amsmath}
\usepackage{amssymb}
\usepackage{pifont}
\usepackage{bm}
\usepackage[most]{tcolorbox}
\usepackage{enumitem}

\newcommand{\cmark}{\textcolor{green!60!black}{\ding{51}}}
\newcommand{\xmark}{\textcolor{red!70!black}{\ding{55}}}

\definecolor{headercolor}{RGB}{230, 230, 230}
\definecolor{highlightcolor}{HTML}{D0E8FF}
\definecolor{altcolor}{RGB}{245, 245, 245}

%% file: sec/1_intro.tex
\section{Introduction}

Recent advances in video generation and editing have significantly expanded the possibilities of content creation. However, a critical gap remains in cinematic background replacement: existing systems often fail to generate backgrounds that move consistently with the foreground’s cinematographic motion. In practice, when the foreground involves professional camera movements (e.g., push, pan, tracking shots), synthesized backgrounds frequently appear static, misaligned, or visibly composited. Crucially, most existing methods implicitly rely on the unrealistic assumption of perfect, pixel-accurate foreground mattes. In real-world inference, handling imperfect segmentation is inevitable; flickering edges and broken contours amplify artifacts and reduce realism. Furthermore, methods attempting global harmonization often struggle to maintain the subject's identity. 

Through systematic analysis, we identify that high-quality cinematic background synthesis requires four key properties: \textbf{motion consistency} (explicitly modeling implied camera movement and dynamic trajectory), \textbf{illumination consistency} (scene-adaptive foreground relighting), \textbf{boundary robustness} (tolerating imperfect masks and maintaining coherent transitions), and \textbf{foreground consistency} (preserving subject identity and fine textures).

To achieve this, we introduce \textbf{PAI-Studio}, a novel video synthesis framework for multi-condition cinematic background replacement. Given a foreground video and one to three background reference images, PAI-Studio generates a complete video satisfying all four properties. We explicitly support \textbf{multi-frame background control} to handle significant background content variations under large camera motion, aligning reference images with designated output frames through positional encoding cloning. At the model level, we build upon an in-context Diffusion Transformer~\cite{dit, song2025omniconsistency, liu2025omnipsd, zhang2025easycontrol, song2025mitty, song2026omnihumanoid, song2026vista} backbone with native bidirectional self-attention. Unlike existing pipelines that stack complex cascaded modules (e.g., dedicated matting networks or relighting adapters), our approach is remarkably elegant and globally reasons over multi-modal conditions without architectural bloat. This \emph{global video generation} strategy rather than local background inpainting naturally reconstructs plausible transitions even with incomplete mask edges, thereby achieving \textbf{boundary robustness}.

To empower this framework, we construct \textbf{CineStudio}, a 30K-scale dataset explicitly reverse-engineered from real-world user inference scenarios. Crucially, we observe that large-scale film and online video data inherently exhibit causal consistency between foreground motion and background evolution under real camera movements, providing natural supervision for learning motion-aligned background synthesis. Therefore, derived from high-quality films and online videos, we perform shot segmentation and optical-flow-based filtering to retain sequences with significant camera and background motion, which explicitly enhances the \textbf{motion consistency} of the synthesized results. Furthermore, to enforce \textbf{illumination consistency}, our data curation augments foreground videos with controlled relighting perturbations. The resulting dataset contains diverse cinematographic movements, scene layouts, and lighting conditions, providing a realistic foundation for studying motion-aware background synthesis.

Compared to existing methods (Table~\ref{tab:feature}), our approach provides a more comprehensive solution. Our main contributions are threefold:
\begin{itemize}
    \item We introduce \textbf{PAI-Studio}, a comprehensive video synthesis framework for cinematic background replacement that explicitly requires motion consistency, illumination consistency, and boundary robustness, while supporting multi-frame background control.
    \item We propose an in-context Diffusion Transformer architecture with bidirectional attention and temporally aligned positional encoding, combining data-driven learning and structured video understanding to achieve motion-aligned background generation and coherent relighting without complex cascaded modules.
    \item We construct \textbf{CineStudio}, a 30K-scale dataset from films and online videos, whose construction is reverse-engineered from inference challenges to support the systematic evaluation of motion-aware background generation.
\end{itemize}

\begin{table*}[!ht]
\centering
\caption{Feature comparison between our method and existing approaches. VACE* denotes our reproduction trained on the same dataset.}
\label{tab:feature}
\rowcolors{3}{altcolor}{white}
\resizebox{\linewidth}{!}{\footnotesize
\begin{tabular}{l||c|c|c|c|c|c|c}
\hline\hline
\rowcolor{headercolor}
\textbf{Method} & \textbf{1080P} & \shortstack{\textbf{Multi-frame}\\\textbf{BG Control}} & \shortstack{\textbf{Moving}\\\textbf{Background}} & \shortstack{\textbf{Foreground}\\\textbf{Consistency}} & \shortstack{\textbf{Illumination}\\\textbf{Harmony}} & \shortstack{\textbf{Edge}\\\textbf{Harmony}} & \shortstack{\textbf{Background}\\\textbf{Consistency}} \\
\hline\hline
VACE* & \cmark & \xmark & \cmark & \xmark & \cmark & \cmark & \xmark \\
Kling O1 & \cmark & \cmark & \cmark & \xmark & \cmark & \cmark & \cmark \\
Runway Gen-4 Aleph & \cmark & \xmark & \xmark & \xmark & \xmark & \xmark & \xmark \\
Kling 3.0 Omni & \xmark & \cmark & \cmark & \xmark & \cmark & \cmark & \xmark \\
Beeble AI Switchx & \cmark & \xmark & \xmark & \cmark & \xmark & \cmark & \cmark \\
\hline
\rowcolor{highlightcolor}
\textbf{Ours} & \cmark & \cmark & \cmark & \cmark & \cmark & \cmark & \cmark \\
\hline\hline
\end{tabular}}
\end{table*}

%% file: sec/2_related.tex
\section{Related Works}
\label{sec:related}

\subsection{Video Diffusion Models}
Video diffusion models have rapidly evolved from 2D U-Nets augmented with temporal layers~\cite{videoldm, videocrafter, gen1} to native spatiotemporal Diffusion Transformers (DiTs)~\cite{dit, sparsevdit}. DiTs treat video data as patch sequences to capture long-range dependencies, and recent large-scale models~\cite{lumina-video, brooks2024video, wan2025, cogvideox, hunyuanvideo} have demonstrated exceptional scalability and high-fidelity generation. This unified architecture provides the structural foundation for our framework, enabling scalable integration of diverse latent representations.

\subsection{Controllable Video Generation}

Controllable video generation aims to synthesize temporally coherent videos under various conditionings, progressing from text-to-video~\cite{imagenvideo, videocrafter, cogvideo, lumina-video} to richer spatial controls~\cite{videocontrolnet, poseguidedvideo, ditctrl, ma2024followyouremoji, ma2025followyourclick, ma2025followyourmotion}. For appearance and motion guidance, image-conditioned approaches~\cite{videoldm, dynamiCrafter, motioncanvas} often struggle with large camera motions, while video-conditioned methods~\cite{vid2vid, makeavideo, gen1, song2025worldwander} require full reference videos, limiting flexibility. Recent advances also explore process-level supervision~\cite{processpainter, song2025makeanything, song2025layertracer}. Despite these efforts, existing methods typically treat backgrounds as static or weakly dynamic and fail to explicitly model camera-aware background motion conditioned on foreground dynamics. In contrast, our work focuses on reference-conditioned background synthesis with explicit temporal alignment and multi-frame background control.

\subsection{Video Editing and Background Replacement}

Classical video background replacement relies on matting and compositing~\cite{bgmattingv2, videomatte, robustvideomatting, chen2025transanimate}, which requires prepared backgrounds and struggles with moving cameras. Recent diffusion-based video editing methods excel at inpainting~\cite{propainter, e2fgvi} and text-driven modifications~\cite{fatezero, tuneavideo, rerenderavideo, text2live, tokenflow, videditdiff, moca}, but typically assume static backgrounds or lack global camera motion modeling. While some works combine matting with generative models~\cite{layerdiffusion, bachvid}, they often synthesize backgrounds independently of foreground motion, causing temporal inconsistencies. Our method directly addresses this by enforcing temporal alignment via bidirectional attention and position-aware conditioning, enabling camera-aware motion and scene-consistent illumination.

%% file: sec/3_method.tex
\section{Method}

\begin{figure*}[!t]
\centering
\includegraphics[width=\linewidth]{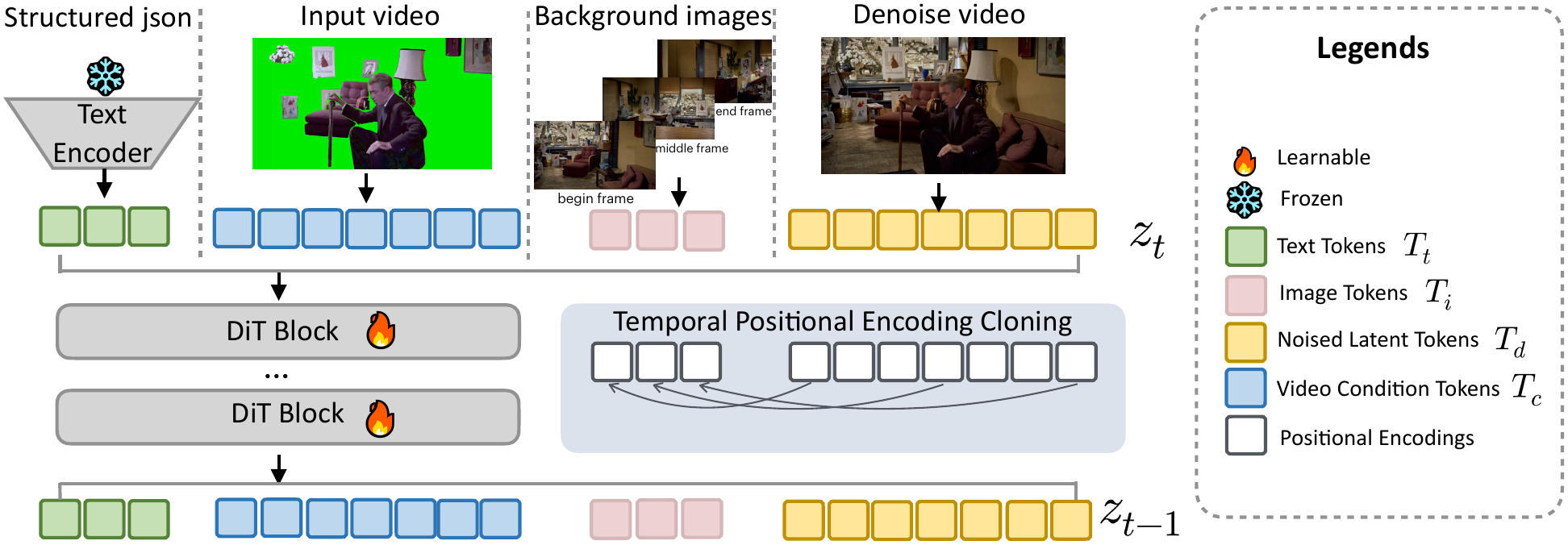}
\caption{Overview of the PAI-Studio architecture. Multi-condition inputs—including multiple background reference images, the illumination-perturbed foreground video, a structured prompt JSON, and the denoising video—are encoded into tokens and concatenated. A multi-modal attention model leverages bidirectional attention to model their global correlations. In addition, Temporal Positional Encoding (PE) Cloning is introduced to precisely control the temporal placement of multiple background images in the generated video.}
\label{fig:method}
\end{figure*}

\subsection{Problem Formulation and Inputs}
\label{sec:problem}

We study cinematic background replacement as a multi-condition video synthesis task.
Given a foreground video $V^{fg}=\{v^{fg}_t\}_{t=1}^{T}$ and a small set of background reference images $\{I^{bg}_k\}_{k=1}^{K}$ with $K\in\{1,2,3\}$, the goal is to synthesize an output video $V^{out}=\{v^{out}_t\}_{t=1}^{T}$ that satisfies the following requirements:

\textbf{(1) Background appearance consistency.}
The synthesized background should match the reference scene appearance indicated by $\{I^{bg}_k\}$.
We explicitly support \textbf{multi-frame background control} since under large camera motion, the visible background content can vary substantially over time, making single-frame reference insufficient.

\textbf{(2) Motion consistency.}
The background dynamics in $V^{out}$ should be consistent with the cinematographic motion implied by the foreground video $V^{fg}$, producing coherent camera-aware background motion rather than static or drifting backgrounds.

\textbf{(3) Foreground fidelity with scene-adaptive relighting.}
The foreground content in $V^{out}$ should preserve the identity and fine details of $V^{fg}$, while its appearance (e.g., shading, color tone) should be adaptively adjusted to better match the target scene implied by $\{I^{bg}_k\}$, i.e., realistic foreground relighting and global illumination harmonization.

\textbf{(4) Robustness to imperfect foreground boundaries.}
In practice, foreground masks from video segmentation often exhibit temporal instability (e.g., flickering, missing parts that reappear in later frames).
$V^{out}$ should remain visually coherent under such imperfections, with plausible boundary completion and temporal consistency enabled by spatiotemporal context.

Optionally, a text prompt $c^{txt}$ can be provided to describe the target scene. We denote all conditions as
$\mathcal{C}=\big(V^{fg}, \{I^{bg}_k\}_{k=1}^{K}, c^{txt}\big)$.

\subsection{Overall Architecture}
\label{sec:architecture}

Our approach builds upon a Diffusion Transformer (DiT) video generation backbone (WAN~2.2-5B) \cite{wan2025}, adapted to a conditional video-to-video synthesis setting with multiple heterogeneous conditions.
All videos and images are first encoded into a shared latent space using a pretrained video VAE, producing latent tokens for the foreground video, background reference images, and text. An overview of the proposed architecture is illustrated in Fig.~\ref{fig:method}.

At each diffusion step, we form a unified token sequence and process it with a single-stream Transformer using \textbf{bidirectional self-attention}, enabling joint reasoning over foreground dynamics, reference scene appearance, and the evolving denoising state. The denoised latent is decoded back into $V^{out}$ by the VAE decoder. Coupled with our tailored dataset, this native bidirectional attention mechanism efficiently resolves multi-modal consistency and global harmonization challenges.

\paragraph{Full bidirectional self-attention.}
Let $Z^{txt}$ denote the text tokens (optional), $Z^{fg}$ the foreground video tokens, $\{Z^{bg}_k\}_{k=1}^{K}$ the background reference tokens, and $Z_s$ the noisy output video tokens at diffusion step $s$.
We concatenate them into a single sequence:
\begin{equation}
X = \big[\, Z^{txt}\ \Vert\ Z^{fg}\ \Vert\ Z^{bg}_1\ \Vert\ \cdots\ \Vert\ Z^{bg}_K\ \Vert\ Z_s \,\big]\in\mathbb{R}^{N\times d},
\label{eq:token_concat}
\end{equation}
where $N$ is the total number of tokens and $d$ is the token dimension.
We apply \emph{full} (bidirectional) self-attention on $X$. For the $h$-th head,
\begin{equation}
Q_h = XW_h^{Q},\quad K_h = XW_h^{K},\quad V_h = XW_h^{V},
\end{equation}
\begin{equation}
\mathrm{Attn}_h(X) = \mathrm{softmax}\!\left(\frac{Q_hK_h^{\top}}{\sqrt{d_h}}\right)V_h,
\label{eq:full_attn}
\end{equation}
and the multi-head output is
\begin{equation}
\mathrm{MHSA}(X)=\mathrm{Concat}\big(\mathrm{Attn}_1(X),\ldots,\mathrm{Attn}_H(X)\big)W^{O},
\label{eq:mhsa}
\end{equation}
where $H$ is the number of heads and $d_h=d/H$.

\paragraph{Structured motion understanding.}
To strengthen motion and camera-awareness, we leverage a strong video understanding model (Gemini-3) \cite{gemini} to extract structured cues from videos, including foreground/background motion summaries, camera movement descriptions, and semantic layer descriptions.
These structured signals are used for dataset curation and can also be provided at inference time as high-level guidance derived from the input foreground video, improving the stability of motion-aligned synthesis.

\begin{figure*}[!t]
\centering
\includegraphics[width=\linewidth]{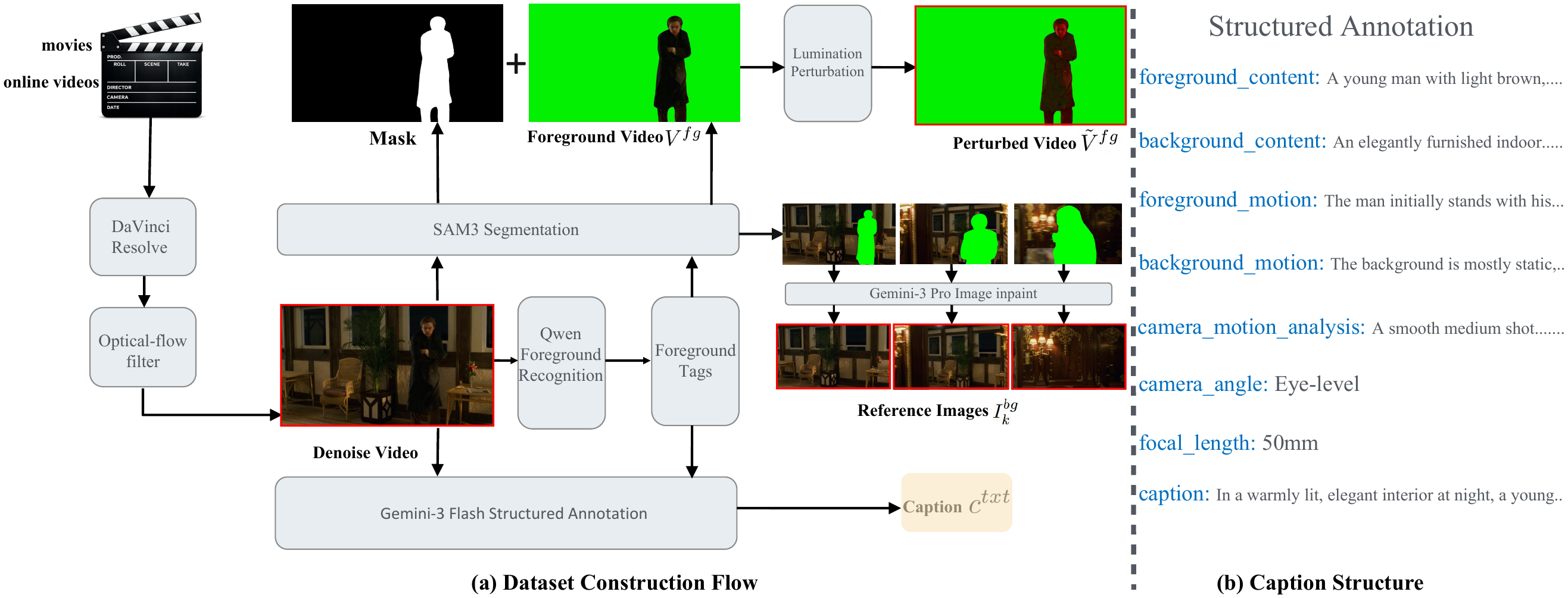}
\caption{Overview of the CineStudio data construction pipeline.}
\label{fig:data_pipeline}
\end{figure*}

\subsection{Multi-Frame Background Control via Temporal Positional Encoding Cloning}
\label{sec:pe}

Single-frame background conditioning is often insufficient under large cinematographic motion, since the visible background content can change substantially across time.
We therefore support \textbf{multi-frame background control} with $K\in\{1,2,3\}$ reference images sampled from designated temporal locations $\{t_k\}_{k=1}^{K}\subset\{1,\ldots,T\}$ (e.g., first/middle/last).

A key challenge is to establish \emph{explicit temporal correspondence} between each background reference $I^{bg}_k$ and a specific output time $t_k$.
We address this with \textbf{temporal positional encoding cloning}.
Let $\mathbf{p}(t)$ denote the temporal positional encoding used by the DiT for time index $t$.
For each background reference token $Z^{bg}_k$, we \emph{clone} the temporal positional encoding from the denoising tokens at time $t_k$:
\begin{equation}
\mathbf{p}\big(Z^{bg}_k\big) \leftarrow \mathbf{p}(t_k), \quad k=1,\ldots,K.
\label{eq:pe_clone_bg}
\end{equation}
This design anchors each reference image to a designated output frame and enables controllable, time-aware conditioning.

Moreover, to enforce motion consistency between the input foreground and the synthesized video, we align the temporal positional encodings of corresponding foreground and denoising frames:
\begin{equation}
\mathbf{p}\big(Z^{fg}_t\big) \equiv \mathbf{p}\big(Z_s(t)\big), \quad t=1,\ldots,T,
\label{eq:pe_align_fg}
\end{equation}
where $Z_s(t)$ denotes the denoising tokens associated with output time $t$.
Together, Eqs.~(\ref{eq:pe_clone_bg})--(\ref{eq:pe_align_fg}) provide a simple yet effective mechanism for \emph{temporal controllability} and motion-aligned background synthesis under multi-frame conditioning.

\subsection{Illumination Perturbation for Scene-Consistent Relighting}
\label{sec:illum}

Achieving illumination consistency requires the model to infer foreground lighting from the target scene rather than copying the input foreground illumination.
We adopt a simple but effective strategy during data construction: we intentionally apply \textbf{relighting and color perturbations} to the foreground video to form $\tilde{V}^{fg}$.
This breaks the direct illumination correspondence between the input foreground and the target video, and forces the model to recover appropriate foreground lighting from background/environment cues.

Concretely, the perturbation operator $\mathcal{R}(\cdot)$ applies global photometric augmentations, including HSV-channel jittering, contrast adjustment, color temperature shifts, and gamma correction:
\begin{equation}
\tilde{V}^{fg} = \mathcal{R}\!\left(V^{fg}\right).
\label{eq:relight_aug}
\end{equation}
We then train the model with $\tilde{V}^{fg}$ as the conditioning foreground while keeping $V^{out}$ as the target.
As a result, the model learns to produce foreground appearances that are consistent with the reference scene illumination, enabling coherent relighting at inference time.


\begin{figure*}[!t]
\animategraphics[width=\linewidth]{5}{demo_v10_frames/demo_}{00}{09}
\caption{Generation results of PAI-Studio. Readers can click and play the video clips in this figure using {\color{red}\textbf{Adobe Acrobat}}.}
\label{fig:results}
\end{figure*}

\subsection{Dataset Construction}
\label{sec:data}

The core philosophy of our data construction pipeline is to explicitly reverse-engineer the training data based on the challenges encountered in real-world user inference. In practical virtual studio scenarios, users consistently struggle with mismatched lighting, imperfect green-screen matting, and the need for temporal coherence. Therefore, rather than collecting ideal synthetic data, we construct our 30K-scale dataset, \textbf{CineStudio}, to directly reflect these difficulties: we purposefully inject controlled relighting perturbations, and leverage the natural cinematographic motion of real-world videos. Derived from high-quality films and online videos, each sample contains $(\tilde{V}^{fg}, \{I^{bg}_k\}_{k=1}^{K}, c^{txt}, V^{out})$, where $V^{out}$ is the original video clip. The overall data construction process is illustrated in Fig.~\ref{fig:data_pipeline}.

\paragraph{Shot segmentation and motion filtering.}
We first perform shot segmentation and compute optical-flow statistics to filter out clips with weak camera/background motion, retaining sequences with clear cinematographic dynamics.


\paragraph{Foreground--background separation and alignment.}
We perform foreground recognition using Qwen3-VL~\cite{qwen3} to obtain a short tag describing the dominant foreground concept. This tag is fed to an advanced video segmentation model (SAM3)~\cite{sam3} to segment and track the foreground over time, producing temporally aligned masks. These masks are then used to extract the foreground onto a green screen, yielding the final foreground video.
The resulting masks may still contain realistic imperfections such as boundary flickering, small holes, or missing regions.

\paragraph{Multi-frame background references.}
We sample $K\in\{1,2,3\}$ background reference frames from different temporal locations (e.g., begin/middle/end), remove the foreground content, and employ Gemini-3 Pro Image~\cite{gemini} to inpaint the missing regions, yielding clean background references $\{I^{bg}_k\}$.

\paragraph{Foreground relighting augmentation.}
We generate relit variants $\tilde{V}^{fg}$ using the perturbation operator $\mathcal{R}(\cdot)$ (Eq.~\eqref{eq:relight_aug}) to encourage scene-consistent relighting. This augmentation explicitly simulates the inference-time challenge where foreground subjects must dynamically adapt to entirely novel and unpredictable background lighting conditions.

\paragraph{Structured annotation with Gemini-3.}
Finally, we use Gemini-3 \cite{gemini} to generate structured JSON annotations for each clip, including foreground/background semantic descriptions, foreground/background motion, camera motion analysis, as well as camera angle and focal length. We also produce a concise caption that summarizes and unifies these fields into a single natural-language description.
These annotations are used for dataset curation and for providing high-level motion/camera guidance at inference time.

%% file: sec/4_experiment.tex
\section{Experiments}
\label{sec:experiments}

\begin{figure*}[!t]
\animategraphics[width=\linewidth]{5}{comparison_v9_frames/comparison_}{00}{09}
\caption{Comparison results show that our method consistently outperforms baseline approaches in structural fidelity, illumination harmony, and temporal coherence. Readers can click and play the video clips in this figure using {\color{red}\textbf{Adobe Acrobat}}.}
\label{fig:comp2}
\end{figure*}

\subsection{Experimental Setup}
\label{sec:setup}

\paragraph{\textbf{Setting.}}
We evaluate on the proposed CineStudio dataset (approximately 30K clips with diverse motions, layouts, and lighting). Each clip contains a foreground video, a target video, and 1-3 background references (defaulting to the \textbf{begin, middle, and end} source frames). Videos are generated at 1080P ($1920 \times 1056$), 24 FPS, and 97 frames (${\sim}4$s). We finetune via LoRA~\cite{hu2022lora} (rank 80) and AdamW~\cite{loshchilov2019adamw} (LR $5\times10^{-5}$) in BF16. Training takes approximately 6,000 steps on 8 H200 GPUs (batch size 1/GPU). Baselines use optimal settings.

\paragraph{\textbf{Baseline Methods.}}
We compare our method with both open-source and commercial video editing systems
that support background replacement or scene synthesis.

(1) Video-editing Baseline. To ensure fair comparison under the same setting, we first exclude models lacking IV2V support. While we surveyed numerous IV2V models, most focus on style transfer or foreground object editing and perform poorly on green-screen background replacement, making them unsuitable as direct baselines (due to space limits, visual results of these baselines are provided in the \textbf{supplementary material}). Therefore, we solely adopt VACE~\cite{vace2025} as our open-source baseline. We finetune the VACE-1.3B model on our CineStudio dataset using the same training data and resolution settings as our method.

(2) Commercial APIs.
We evaluate Kling O1, Kling 3.0 Omni, Runway Gen-4 Aleph (denoted as Runway Aleph), and Beeble AI Switchx (denoted as Switchx),
which provide background replacement or generative scene synthesis functionality.
These systems are evaluated under their default API configurations.

\paragraph{\textbf{Benchmark.}}
To rigorously evaluate the proposed method, we introduce a comprehensive benchmark, \textbf{Cine-Benchmark}, structured into two distinct subsets to decouple the evaluation of context fidelity and out-of-distribution generalization:

(1) Cine-Restore Subset focuses on \emph{Existing-Frame Background Replacement}. It consists of 150 videos randomly sampled from our \textbf{CineStudio} dataset (disjoint from training). In this subset, the background reference frames are clean plates extracted and inpainted from the original source video. This simulates a closed-loop background completion task where the model must faithfully restore the original cinematic context under new lighting conditions. For this subset, we compute quantitative metrics by directly comparing the generated results against the paired ground-truth videos.

(2) Cine-NBG (Novel Background Generation) Subset evaluates the model's synthesis and generalization capabilities in entirely new environments. This \emph{Novel Background Replacement} subset comprises 50 videos curated from public sources (DAVIS 2017~\cite{davis2017} and IVEBench~\cite{chen2025ivebench}). Unlike Cine-Restore, the background reference frames in Cine-NBG are synthesized by modifying the original background using Gemini to create diverse, out-of-distribution styles and layouts. This subset tests the model's robustness under real-world camera motions and its ability to generate plausible new backgrounds. Since no ground-truth exists for these newly generated scenes, we evaluate performance using Gemini-based perceptual scores.

\paragraph{\textbf{Evaluation Metrics.}}
Following the objectives in Sec.~\ref{sec:problem}, we evaluate our model across four dimensions (see the \textbf{supplementary material} for Gemini scoring details): 

(1) Background Appearance Consistency, evaluated by the average per-frame \textbf{Mean Squared Error (MSE)}, \textbf{Structural Similarity (SSIM)}~\cite{ssim}, and \textbf{Learned Perceptual Image Patch Similarity (LPIPS)}~\cite{lpips} between ground truth and output videos to assess restored scene fidelity;

(2) Motion Consistency (\textbf{Motion Cons.}), evaluated via CoTracker3~\cite{cotracker3} through motion trajectory comparison between ground truth
and output videos for the Cine-Restore subset, and via Gemini-based perceptual scoring for the Cine-NBG subset, ensuring background dynamics strictly follow foreground camera motion;

(3) Foreground Fidelity and Relighting, evaluated via Gemini-based perceptual scoring, using \textbf{Foreground Preservation (Fg. Pres.)} to quantify foreground identity preservation, and \textbf{Illumination Harmony (Illum. Harmony)} to assess the effectiveness of scene-adaptive relighting;

(4) Seamless Integration, also evaluated via Gemini-based perceptual scoring, using \textbf{Edge Quality} to assess the model's ability to handle boundary imperfections, and \textbf{Foreground--Background Fusion (FG-BG Fusion)} to quantify the global geometric, photometric, and temporal coherence at the foreground--background interface, ensuring a visually seamless composition.

\begin{table*}[!t]
\centering
\caption{
Quantitative comparison on the Cine-Restore subset under both single-frame and multi-frame settings. The last three metrics are evaluated via Gemini-based perceptual scoring \textbf{(scale 1-10)}, whereas the rest are standard algorithmic metrics. Multi-frame results are reported only for methods that support multi-frame control.
}
\label{tab:quantitative}
\rowcolors{3}{altcolor}{white} 
\resizebox{\textwidth}{!}{\footnotesize
\begin{tabular}{l||cc|cc|cc|cc|cc|cc|cc}
\hline\hline
\rowcolor{headercolor}
& \multicolumn{2}{c|}{\textbf{MSE} $\downarrow$}
& \multicolumn{2}{c|}{\textbf{SSIM} $\uparrow$}
& \multicolumn{2}{c|}{\textbf{LPIPS} $\downarrow$}
& \multicolumn{2}{c|}{\textbf{Motion Cons.} $\uparrow$}
& \multicolumn{2}{c|}{\textbf{Edge Quality} $\uparrow$}
& \multicolumn{2}{c|}{\textbf{FG-BG Fusion} $\uparrow$}
& \multicolumn{2}{c}{\textbf{Illum. Harmony} $\uparrow$} \\
\cline{2-15}
\rowcolor{headercolor}
\multirow{-2}{*}{\textbf{Method}}
& \textbf{Single} & \textbf{Multi}
& \textbf{Single} & \textbf{Multi}
& \textbf{Single} & \textbf{Multi}
& \textbf{Single} & \textbf{Multi}
& \textbf{Single} & \textbf{Multi}
& \textbf{Single} & \textbf{Multi}
& \textbf{Single} & \textbf{Multi} \\
\hline\hline 

VACE (Open-source) & 0.0270 & -- & 0.595 & -- & 0.393 & -- & 0.792 & -- & 7.19 & -- & 7.33 & -- & 7.67 & -- \\
Kling O1 (Commercial) & 0.0282 & 0.0190 & 0.646 & 0.682 & 0.402 & 0.341 & 0.706 & 0.721 & 8.44 & 9.04 & 7.88 & 8.20 & 7.62 & 7.44 \\
Kling 3.0 Omni (Commercial) & 0.0288 & 0.0210 & 0.592 & 0.622 & 0.414 & 0.370 & 0.787 & 0.715 & 8.12 & 8.40 & 7.68 & 7.96 & 7.92 & 7.44 \\
Runway Aleph (Commercial) & 0.0692 & -- & 0.413 & -- & 0.657 & -- & 0.323 & -- & 5.54 & -- & 4.38 & -- & 5.88 & -- \\
Switchx (Commercial) & 0.0320 & -- & 0.668 & -- & 0.415 & -- & 0.676 & -- & 7.08 & -- & 6.36 & -- & 7.00 & -- \\
\hline 
\rowcolor{highlightcolor}
\textbf{Ours} & \textbf{0.0164} & \textbf{0.00645} & \textbf{0.717} & \textbf{0.789} & \textbf{0.303} & \textbf{0.197} & \textbf{0.872} & \textbf{0.894} & \textbf{9.17} & \textbf{9.62} & \textbf{8.46} & \textbf{9.33} & \textbf{9.11} & \textbf{9.61} \\
\hline\hline 
\end{tabular}}
\end{table*}

\begin{table*}[!t]
\centering
\caption{Gemini-based perceptual evaluation on the Cine-NBG subset. Only Gemini-based scores \textbf{(scale 1-10)} are reported as no ground-truth video is available.}
\label{tab:gemini}
\rowcolors{2}{altcolor}{white} 
\resizebox{\textwidth}{!}{
\begin{tabular}{l||c|c|c|c|c}
\hline\hline
\rowcolor{headercolor} 
\textbf{Method} 
& \textbf{Motion Cons.} $\uparrow$ 
& \textbf{Edge Quality} $\uparrow$ 
& \textbf{Fg. Pres.} $\uparrow$ 
& \textbf{FG-BG Fusion} $\uparrow$ 
& \textbf{Illum. Harmony} $\uparrow$ \\
\hline\hline

VACE (Open-source) & 6.92 & 7.82 & 2.44 & 7.04 & 8.38 \\
Kling O1 (Commercial) & 7.12 & 8.04 & 6.48 & 7.36 & 7.72 \\
Runway Aleph (Commercial) & 6.21 & 7.21 & 6.38 & 6.29 & 7.29 \\
Kling 3.0 Omni (Commercial) & 7.32 & 7.80 & 5.96 & 6.76 & 8.52 \\
Switchx (Commercial) & 7.12 & 7.40 & 8.88 & 6.16 & 7.88 \\
\hline
\rowcolor{highlightcolor}
\textbf{Ours} & \textbf{8.16} & \textbf{8.86} & \textbf{9.58} & \textbf{8.40} & \textbf{9.22} \\
\hline\hline
\end{tabular}}
\end{table*}

\subsection{Results}
\label{sec:results}

The generation results are shown in Figure~\ref{fig:results}. A key advantage of our approach is its strong temporal stability across long sequences. Unlike existing editing-based methods that suffer from flickering or static background artifacts, our model produces camera-aware background motion that closely follows foreground cinematographic dynamics, significantly reducing compositing artifacts even under minor matting imperfections. Furthermore, our framework achieves scene-consistent foreground relighting. When background references exhibit distinct lighting conditions—such as strong directional illumination or low-light environments—the synthesized videos maintain globally coherent shading and color adaptation on foreground subjects, preventing visual detachment between layers. Crucially, despite the global generative formulation, foreground identity and fine-grained textures are rigorously preserved across frames, avoiding the structural distortions and identity drift commonly observed in diffusion-based baselines. Finally, multi-reference background conditioning proves essential in large-motion scenarios. By aligning references from different temporal locations via positional encoding cloning, the model maintains appearance consistency under substantial viewpoint changes. Overall, explicitly modeling camera-aware motion and scene-consistent illumination enables high-quality cinematic background synthesis beyond existing approaches. More video results on edge quality and robustness to imperfect foregrounds are in the \textbf{supplementary material}.

\begin{figure*}[!t]
\animategraphics[width=\linewidth]{5}{ablation_v3_frames_v5/ablation_}{00}{09}
\caption{\textbf{Ablation study results.} The full model achieves the best or near-best performance, validating the importance of position encoding for motion consistency, illumination adaptation for realistic integration, and structured annotation for detailed guidance. Readers can click and play the videos in this figure using {\color{red}\textbf{Adobe Acrobat}}.}
\label{fig:ablation_vis}
\end{figure*}

\begin{table*}[!t]
\centering
\caption{Ablation study on the Cine-Restore subset. We evaluate each component under both single-frame and multi-frame control protocols.}
\label{tab:ablation}
\rowcolors{3}{altcolor}{white} 
\resizebox{\linewidth}{!}{\footnotesize
\begin{tabular}{l||cc|cc|cc|cc|cc|cc|cc}
\hline\hline
\rowcolor{headercolor}
& \multicolumn{2}{c|}{\textbf{MSE} $\downarrow$}
& \multicolumn{2}{c|}{\textbf{SSIM} $\uparrow$}
& \multicolumn{2}{c|}{\textbf{LPIPS} $\downarrow$}
& \multicolumn{2}{c|}{\textbf{Motion Cons.} $\uparrow$}
& \multicolumn{2}{c|}{\textbf{Edge Quality} $\uparrow$}
& \multicolumn{2}{c|}{\textbf{FG-BG Fusion} $\uparrow$}
& \multicolumn{2}{c}{\textbf{Illum. Harmony} $\uparrow$} \\
\cline{2-15}
\rowcolor{headercolor}
\multirow{-2}{*}{\textbf{Method}}
& \textbf{Single} & \textbf{Multi}
& \textbf{Single} & \textbf{Multi}
& \textbf{Single} & \textbf{Multi}
& \textbf{Single} & \textbf{Multi}
& \textbf{Single} & \textbf{Multi}
& \textbf{Single} & \textbf{Multi}
& \textbf{Single} & \textbf{Multi} \\
\hline\hline

w/o Position Encoding Control
& 0.0300 & 0.0160
& 0.608 & 0.681
& 0.399 & 0.287
& 0.824 & 0.843
& 6.67 & 8.38
& 6.85 & 8.41
& 7.63 & 8.88 \\

w/o Illumination Adaptation
& 0.0270 & 0.0130
& 0.607 & 0.681
& 0.381 & 0.279
& 0.850 & 0.879
& 6.40 & 7.25
& 6.39 & 7.24
& 6.44 & 7.31 \\

w/o Structured Annotation
& 0.0177 & 0.00695
& 0.715 & 0.787
& 0.307 & 0.198
& \textbf{0.873} & \textbf{0.902}
& 9.10 & 9.52
& 8.39 & 9.21
& 8.88 & 9.57 \\

\hline
\rowcolor{highlightcolor}
Full Model
& \textbf{0.0164} & \textbf{0.00645}
& \textbf{0.717} & \textbf{0.789}
& \textbf{0.303} & \textbf{0.197}
& 0.872 & 0.894
& \textbf{9.17} & \textbf{9.62}
& \textbf{8.46} & \textbf{9.33}
& \textbf{9.11} & \textbf{9.61} \\
\hline\hline
\end{tabular}}
\end{table*}

\subsection{Comparison and Evaluation}
\label{sec:comparison}

\paragraph{Quantitative Evaluation.}
Tables~\ref{tab:quantitative} and \ref{tab:gemini} report quantitative comparisons on the Cine-Restore and Cine-NBG subsets. Our method consistently outperforms both open-source baselines and commercial APIs across all evaluation metrics. Specifically, our leading scores in Motion Consistency, alongside reconstruction metrics (MSE, SSIM, and LPIPS), demonstrate that our model generates camera-aware background dynamics while maintaining strict appearance consistency with the reference scene. Notably, multi-frame conditioning via positional encoding cloning (Table~\ref{tab:quantitative}) further improves these results, validating its effectiveness for handling large camera motions. Furthermore, our approach effectively addresses scene-adaptive relighting and imperfect foreground boundaries, as evidenced by top performance in Foreground Preservation (Fg. Pres.), Illumination Harmony, Edge Quality, and FG-BG Fusion. These results confirm that our model can naturally integrate actors into the target scene for both existing background reconstruction (Table~\ref{tab:quantitative}) and new background replacement tasks (Table~\ref{tab:gemini}).

\paragraph{Qualitative Evaluation.}
Figure~\ref{fig:comp2} presents qualitative comparisons.
Existing video editing methods often generate static or weakly dynamic backgrounds, leading to noticeable inconsistencies when the camera moves.
Baseline results typically exhibit misalignment artifacts, inconsistent lighting and noticeable foreground identity distortion, as highlighted in the \textbf{red boxes} in Figure~\ref{fig:comp2}.
In contrast, our method produces backgrounds that evolve coherently over time and exhibit illumination consistent with the target scene.
Crucially, our approach maintains strict foreground consistency, faithfully preserving the actors' identities and structural details without degradation.
The foreground actors appear naturally integrated into the synthesized environment, with reduced compositing artifacts and strong robustness across various scenarios.

\subsection{Ablation Study}
\label{sec:ablation}

We conduct ablation experiments on the Cine-Restore subset to verify our core designs. 
Qualitative results are illustrated in Fig.~\ref{fig:ablation_vis}. As shown in Table~\ref{tab:ablation}, the Full Model achieves the best or near-best performance across all metrics. 
Without Illumination Adaptation, the model fails to automatically adjust foreground lighting to match the target environment, resulting in poor foreground--background integration and notably lower scores in illumination harmony, visually evident from the foreground lighting mismatch with the ground truth in Fig.~\ref{fig:ablation_vis}(a). 
Without Position Encoding Control, although the model produces normal-looking frames, it loses the ability to map background references to precise temporal locations, leading to a significant drop in motion consistency and temporally controllable synthesis, visually evident from the inaccurate camera tilt and road shifting in the top example of Fig.~\ref{fig:ablation_vis}(b). 
Furthermore, relying on simple text prompts rather than our structured annotations degrades quantitative performance: the model misses critical illumination and semantic cues, manifesting as mismatched lighting or background collapse in Fig.~\ref{fig:ablation_vis}(c). Notably, the richer fine-grained captions in our annotations improve robustness even with incomplete keyframe cues, yielding results closer to the ground truth. These results confirm that precise temporal alignment, active relighting, and detailed textual guidance are essential for realistic video synthesis.

\subsection{User Study}
\label{sec:userstudy}

\begin{figure}[tbh]
\centering
\includegraphics[width=1\linewidth]{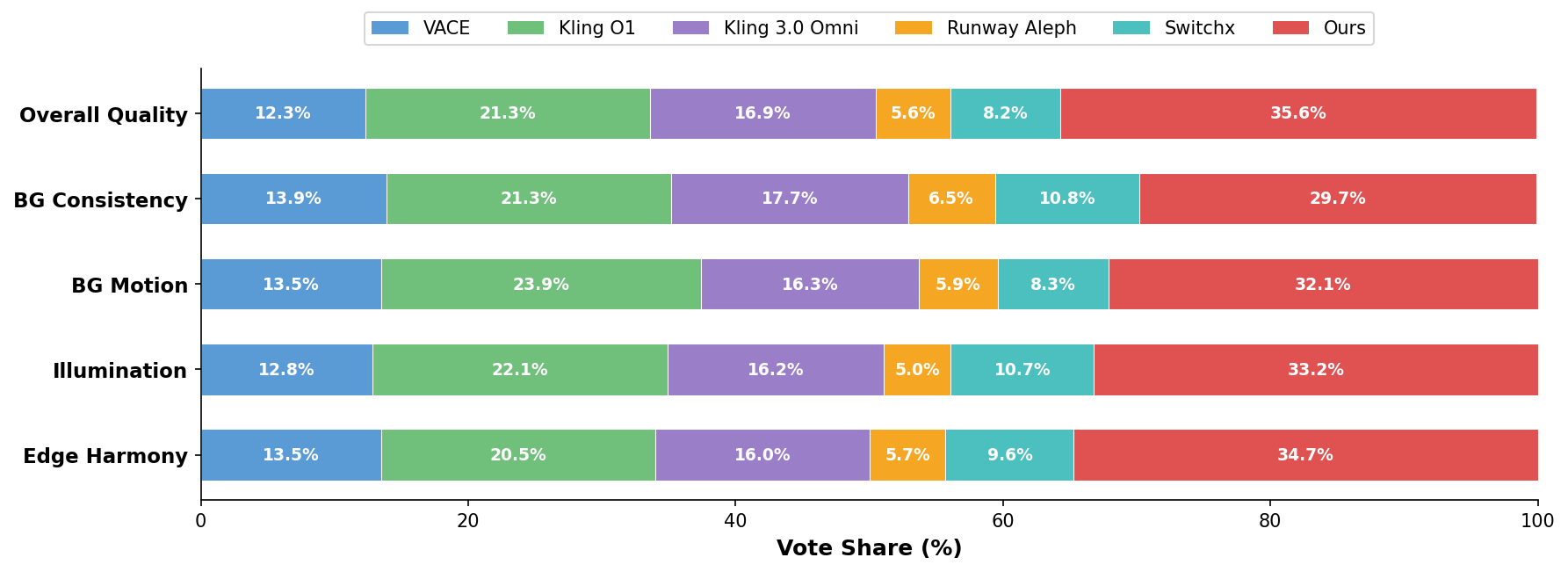}
\caption{User study results comparing our method with open-source methods and commercial APIs. Values indicate the percentage of votes received by each method. Our method is consistently preferred across all five evaluation criteria.}
\label{fig:userstudy}
\end{figure}

We conduct a user study to evaluate perceptual quality and realism from a human perspective, which is particularly important for virtual studio and film production scenarios. We randomly sample 50 videos from both the Cine-Restore and Cine-NBG subsets, and compare our method with representative open-source methods and commercial APIs. A total of 25 participants with prior experience in video editing or visual content creation took part in the study.
Each test case is evaluated by multiple independent participants, and the final scores are reported as the percentage of times each method is voted as the best result. As shown in Figure~\ref{fig:userstudy}, our method is consistently preferred across all five evaluation criteria, aligning with our quantitative evaluations and confirming that our model comprehensively better matches human judgments across all dimensions of the cinematic background replacement task.

%% file: sec/5_conclusion.tex
\section{Conclusion}
\label{sec:conclusion}

We introduced PAI-Studio, a multi-condition video synthesis task for cinematic background replacement that jointly requires motion-aligned background generation, scene-consistent foreground relighting, and robustness to imperfect segmentation boundaries. We presented an in-context Diffusion Transformer framework that concatenates foreground video, multi-frame background references, and denoising tokens into a unified sequence and models them with full bidirectional attention for coherent cross-condition reasoning. Temporal positional encoding cloning enables precise multi-frame temporal control, while foreground relighting and color perturbations during training encourage illumination recovery from scene cues. We further curated a 30K-scale dataset from films and online videos with structured motion annotations. Extensive quantitative and qualitative results demonstrate significant improvements over open-source and commercial API baselines, and user studies confirm a strong preference for our outputs. We believe this work provides a practical foundation toward film-quality controllable video generation and virtual production.

%% file: sec/X_suppl.tex

\setcounter{page}{1}

\twocolumn[{%
  \centering
  {\Large\bfseries PAI-Studio: Cinematic Video Background Replacement with Camera-Aware Motion}\par\vspace{0.5em}
  {\large Supplementary Material}\par\vspace{1.0em}
}]

The supplementary material is organized as follows:
\begin{enumerate}
    \item We provide comprehensive details regarding the user study, including the interface, evaluation criteria, and statistical significance analysis in Sec.~\ref{sec:suppl_user_study}.
    \item We detail the prompts and metrics used for our Gemini-based automated semantic evaluation in Sec.~\ref{sec:suppl_gemini}, and the methodology for evaluating motion fidelity using CoTracker3 in Sec.~\ref{sec:suppl_cotracker}.
    \item We present the statistical correlation between our automated evaluation and human preference in Sec.~\ref{sec:suppl_correlation}.
    \item We analyze the failure modes of the excluded open-source baselines in Sec.~\ref{sec:suppl_bad_baselines}.
    \item We provide additional qualitative results, demonstrating superior edge harmonization (Sec.~\ref{sec:suppl_edge}), robustness to imperfect masks (Sec.~\ref{sec:suppl_robustness}), implicit scene-adaptive relighting (Sec.~\ref{sec:suppl_relighting}), and the effect of multi-frame control (Sec.~\ref{sec:suppl_multiframe}).
    \item We discuss our explorations into efficient conditioning strategies in Sec.~\ref{sec:efficient}.
    \item We provide the prompt templates used for generating structured video annotations in Sec.~\ref{sec:suppl_gemini_caption}.
    \item We report the computational cost and resources for training and inference in Sec.~\ref{sec:suppl_cost}.
    \item We present an analysis of typical failure cases in Sec.~\ref{sec:fail}.
\end{enumerate}

\setcounter{section}{0}
\renewcommand{\thesection}{\Alph{section}}

\section{User Study Details}
\label{sec:suppl_user_study}

\begin{figure}[!t]
    \centering
    \includegraphics[width=\linewidth]{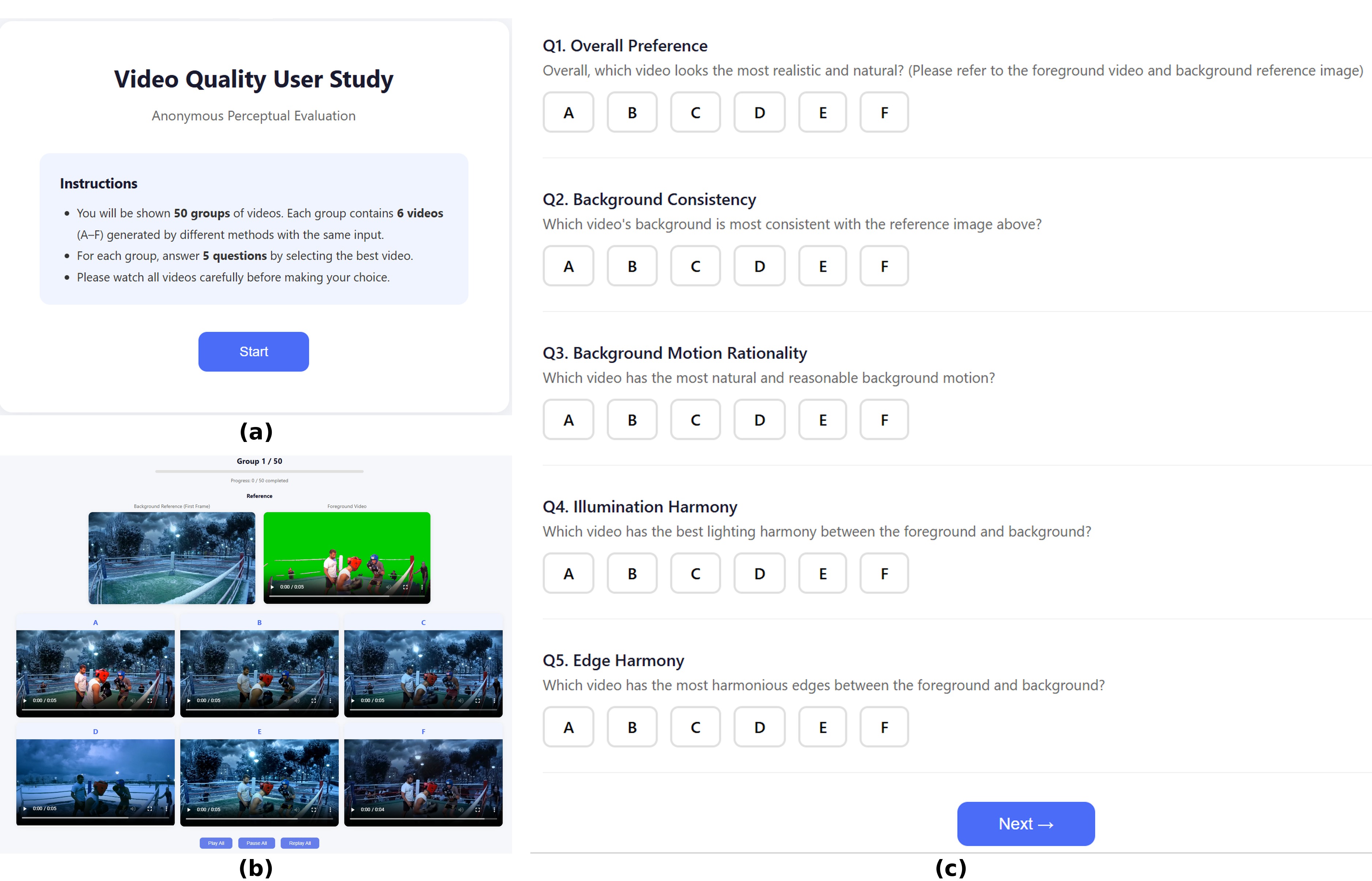}
    \caption{\textbf{User study interface and workflow.} (a) Study instructions, (b) side-by-side presentation of a test case with input references and anonymized results, and (c) the questionnaire with five evaluation criteria (Overall Preference, Background Consistency, Background Motion Rationality, Illumination Harmony, and Edge Harmony).}
    \label{fig:user_study_ui}
\end{figure}

The interface and workflow of our user study are illustrated in Fig.~\ref{fig:user_study_ui}. We designed a comprehensive evaluation pipeline to compare the perceptual quality of our method against five representative baselines.

\textbf{Experimental Protocol.} As shown in the study instructions (Fig.~\ref{fig:user_study_ui}a), participants are required to evaluate 50 groups of videos. For each group, the interface (Fig.~\ref{fig:user_study_ui}b) presents the \textit{Background Reference (First Frame)} and the \textit{Foreground Video} as references at the top. Below these, six anonymized results (A--F) generated under identical inputs are displayed side-by-side. To ensure a fair and thorough assessment, the interface provides synchronized playback controls (Play All, Pause All, and Replay All), allowing participants to scrutinize temporal consistency and motion alignment.

\textbf{Evaluation Criteria.} For each test case, participants must answer five questions by selecting the single best video for each metric (Fig.~\ref{fig:user_study_ui}c):
\begin{enumerate}
\item \textbf{Overall Preference}: Overall, which video looks the most realistic and natural? (Please refer to the foreground video and background reference image.)
\item \textbf{Background Consistency}: Which video's background is most consistent with the reference image?
\item \textbf{Background Motion Rationality}: Which video has the most natural and reasonable background motion?
\item \textbf{Illumination Harmony}: Which video has the best lighting harmony between the foreground and background?
\item \textbf{Edge Harmony}: Which video has the most harmonious edges between the foreground and background?
\end{enumerate}

\textbf{Participants and Scoring.} The study involved 25 independent participants with prior experience in video editing or visual content creation. Participants were instructed to watch all candidate videos carefully before making their choices. The final scores are reported as the percentage of votes each method received for each criterion across all 50 test cases.

\textbf{Randomization and Position Bias.}
To completely eliminate potential position bias, the display order of the six methods was independently and uniformly randomized for each participant and video combination across all 1,250 evaluation trials (25 participants $\times$ 50 videos). A chi-square goodness-of-fit test mathematically confirms that every method appeared at each display position with approximately equal frequency (all $p > 0.1$), indicating the absence of any systematic position bias.

\begin{figure}[!t]
    \centering
    \includegraphics[width=\linewidth]{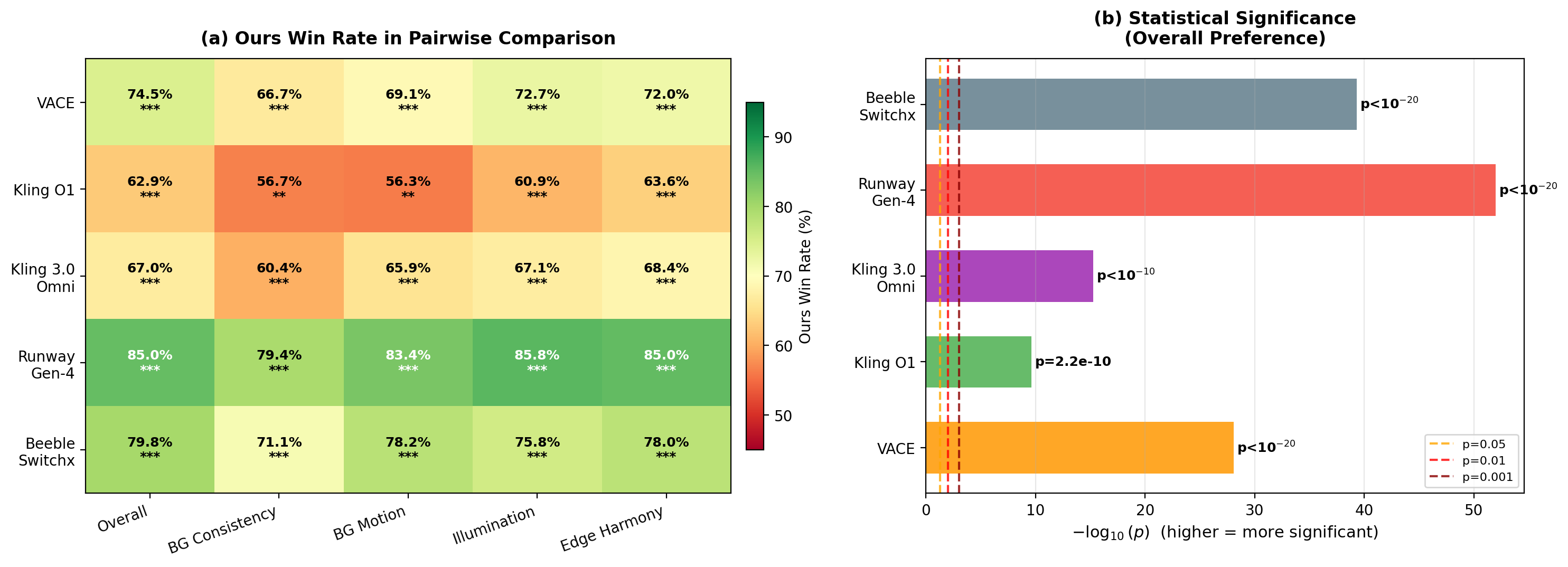}
    \caption{\textbf{Statistical significance of user study results.} (a) Heatmap of our method's win rate (\%) in pairwise comparisons against each baseline across five evaluation dimensions. All cells demonstrate statistical significance ($p < 0.01$, denoted by ** or ***). (b) Significance level ($-\log_{10} p$) for the \textbf{Overall Preference} dimension; all comparisons safely exceed the $p=0.001$ threshold.}
    \label{fig:significance}
\end{figure}
\textbf{Statistical Significance.}
To rigorously validate the reliability of our preference results, we conducted pairwise binomial tests between our method and each baseline. For each comparison, we evaluated the null hypothesis that the preference win rate is merely 50\%. As illustrated in Fig.~\ref{fig:significance}, our method achieves a decisive win rate exceeding 56\% against all baselines across all five evaluation dimensions. Statistical significance is strictly confirmed for every single comparison ($p < 0.01$, with the vast majority reaching $p < 0.001$). Notably, we observe the strongest advantages against Runway Gen-4 and Beeble Switchx (win rates $>75\%$). Furthermore, our method significantly outperforms even the most competitive baseline, Kling O1, securing win rates between 56.3\% and 63.6\%.

\section{Gemini-based Evaluation Details}
\label{sec:suppl_gemini}

We employ Gemini-3-Pro~\cite{gemini} as an automated perceptual judge for semantic-level evaluation. Two complementary protocols are designed: (1)~\textbf{Cine-Restore Subset}, where the generated video is directly compared against the ground-truth (GT) video to evaluate reconstruction fidelity; and (2)~\textbf{Cine-NBG Subset}, where the model assesses the generated video alone (with optional background keyframes). Each metric is queried 3 times and the median score is taken to reduce variance. The full prompt templates are shown in Fig.~\ref{fig:prompt_wgt}, Fig.~\ref{fig:prompt_wogt}, and Fig.~\ref{fig:prompt_wogt_fg}.

\tcbset{
  promptbox/.style={
    colback=blue!3, colframe=blue!50!black,
    fonttitle=\bfseries\small, coltitle=white, colbacktitle=blue!55!black,
    boxrule=0.5pt, arc=2pt, left=4pt, right=4pt, top=2pt, bottom=2pt,
    before skip=4pt, after skip=4pt,
  },
  metricbox/.style={
    colback=gray!5, colframe=gray!50,
    boxrule=0.4pt, arc=1.5pt, left=3pt, right=3pt, top=2pt, bottom=2pt,
    before skip=3pt, after skip=3pt,
  }
}

\begin{figure}[!t]
\begin{tcolorbox}[promptbox, title=Gemini Evaluation Prompts --- Cine-Restore Subset]
\footnotesize

\textbf{Common Setup.}
\textit{Input:} (1)~Generated composited video; (2)~Ground Truth (GT) video as reference.
\textit{Output:} JSON \texttt{\{``score'': <1--10>, ``reason'': ``<sentence>''\}}.

\begin{tcolorbox}[metricbox]
\textbf{Edge Quality} (\texttt{Edge\_Cons}) \hfill \textit{Role: Expert evaluator for matte/edge quality.}\\[2pt]
\textit{Task:} Evaluate edge consistency around the foreground subject, using GT as the reference.\\[1pt]
\textit{Criteria:}
\begin{itemize}[nosep,leftmargin=12pt,label=\textbullet]
\item Halo/fringing, color spill (green/blue edges), jagged edges
\item Temporal edge stability: no crawl/shimmer frame-to-frame (vs.\ GT)
\item Fine structures: hair, fingers, thin objects, semi-transparent regions
\item Holes/missing parts: accidental cut-outs, broken silhouettes
\item Motion blur handling: edges during fast motion should look natural
\end{itemize}
\textit{Scoring:}
10: Clean stable edges matching GT $|$
8--9: Minor noise $|$
6--7: Noticeable artifacts $|$
4--5: Strong issues $|$
1--3: Unusable
\end{tcolorbox}

\begin{tcolorbox}[metricbox]
\textbf{FG--BG Fusion} \hfill \textit{Role: Expert evaluator for compositing realism.}\\[2pt]
\textit{Task:} Evaluate foreground--background integration realism, using GT as reference.\\[1pt]
\textit{Criteria:}
\begin{itemize}[nosep,leftmargin=12pt,label=\textbullet]
\item Scale \& perspective: subject fits the scene geometry (vs.\ GT)
\item Contact \& depth cues: feet contact ground; occlusion boundaries correct
\item Shadows/reflections: consistent with background and subject placement
\item Color \& tone match: foreground grading matches background
\item Temporal integration: no floating, jitter, or background leaking
\end{itemize}
\textit{Scoring:}
10: Seamless, as natural as GT $|$
8--9: Minor mismatch $|$
6--7: Moderate artifacts $|$
4--5: Strong composite feel $|$
1--3: Clearly fake
\end{tcolorbox}

\begin{tcolorbox}[metricbox]
\textbf{Illumination Harmony} (\texttt{Illum\_Harmony}) \hfill \textit{Role: Expert evaluator for lighting consistency.}\\[2pt]
\textit{Task:} Evaluate illumination harmony, using GT as the reference for correct lighting.\\[1pt]
\textit{Criteria:}
\begin{itemize}[nosep,leftmargin=12pt,label=\textbullet]
\item Lighting direction: key light on subject matches scene cues (vs.\ GT)
\item Color temperature: warm/cool balance matches background
\item Intensity/exposure: subject brightness matches scene exposure
\item Shadow plausibility: shadows fit background lighting
\item Temporal consistency: no flickering or unnatural lighting changes
\end{itemize}
\textit{Scoring:}
10: Perfect match, as naturally lit as GT $|$
8--9: Minor mismatch $|$
6--7: Noticeable $|$
4--5: Strong contradiction $|$
1--3: Destroys realism
\end{tcolorbox}

\end{tcolorbox}
\caption{\textbf{Gemini evaluation prompt templates (Cine-Restore Subset).} Three metrics compare the generated composite against the ground truth video. Each prompt instructs Gemini 3 Pro~\cite{gemini} to return a 1--10 score with a brief rationale.}
\label{fig:prompt_wgt}
\end{figure}

\begin{figure}[!t]
\begin{tcolorbox}[promptbox, title=Gemini Evaluation Prompts --- Cine-NBG Subset (Part 1/2)]
\footnotesize

\textbf{Common Setup.}
\textit{Input:} Generated video only (+ optional background keyframes for Motion).
\textit{Output:} JSON \texttt{\{``score'': <1--10>, ``reason'': ``<sentence>''\}}.

\begin{tcolorbox}[metricbox]
\textbf{Motion Consistency} (\texttt{Motion\_Cons}) \hfill \textit{Role: Expert evaluator for temporal realism.}\\[2pt]
\textit{Task:} Evaluate motion consistency \& temporal coherence of the entire video.\\[1pt]
\textit{Criteria:}
\begin{itemize}[nosep,leftmargin=12pt,label=\textbullet]
\item Temporal stability: no flicker, texture crawling, or popping elements
\item Geometry rigidity: static structures remain rigid, no warping/stretching
\item Consistent perspective: lines/edges should not bend frame-to-frame
\item Plausible camera motion: coherent pan/tilt/dolly with consistent parallax
\item Foreground motion: natural, smooth, no jitter or sudden jumps
\end{itemize}
\textit{Scoring:}
10: Stable \& plausible $|$
8--9: Minor flicker $|$
6--7: Noticeable artifacts $|$
4--5: Frequent warping $|$
1--3: Severely unstable
\end{tcolorbox}

\begin{tcolorbox}[metricbox]
\textbf{FG--BG Fusion}\hfill \textit{Role: Expert evaluator for compositing realism.}\\[2pt]
\textit{Task:} Evaluate foreground--background fusion realism.\\[1pt]
\textit{Criteria:}
\begin{itemize}[nosep,leftmargin=12pt,label=\textbullet]
\item Scale \& perspective: subject fits the scene geometry and camera viewpoint
\item Contact \& depth cues: feet contact ground; occlusion boundaries correct
\item Shadows/reflections: consistent with background and subject placement
\item Color \& tone match: foreground grading matches background
\item Temporal integration: no floating, jitter, or background leaking onto subject
\end{itemize}
\textit{Scoring:}
10: Seamless, one real shot $|$
8--9: Minor artifacts $|$
6--7: Moderate $|$
4--5: Strong composite feel $|$
1--3: Clearly fake
\end{tcolorbox}

\begin{tcolorbox}[metricbox]
\textbf{Illumination Harmony} (\texttt{Illum\_Harmony}) \hfill \textit{Role: Expert evaluator for lighting consistency.}\\[2pt]
\textit{Task:} Evaluate illumination harmony between foreground and background.\\[1pt]
\textit{Criteria:}
\begin{itemize}[nosep,leftmargin=12pt,label=\textbullet]
\item Lighting direction: key light direction on subject matches scene cues
\item Color temperature: warm/cool balance matches the background
\item Intensity/exposure: subject brightness matches scene exposure
\item Shadow plausibility: shadows on subject fit background lighting
\item Temporal consistency: lighting should not flicker or change unnaturally
\end{itemize}
\textit{Scoring:}
10: Perfect match $|$
8--9: Minor mismatch $|$
6--7: Noticeable $|$
4--5: Strong contradiction $|$
1--3: Destroys realism
\end{tcolorbox}

\end{tcolorbox}
\caption{\textbf{Gemini evaluation prompt templates (Cine-NBG Subset, part 1/2).} Three metrics assess the generated video directly. Motion\_Cons optionally uses background keyframes. FG--BG\_Fusion and Illum\_Harmony evaluate the video alone without any reference.}
\label{fig:prompt_wogt}
\end{figure}

\begin{figure}[!t]
\begin{tcolorbox}[promptbox, title=Gemini Evaluation Prompts --- Cine-NBG Subset (Part 2/2)]
\footnotesize

\begin{tcolorbox}[metricbox]
\textbf{Edge Quality} (\texttt{Edge\_Cons}) \hfill \textit{Role: Expert evaluator for matte/edge quality.}\\[2pt]
\textit{Task:} Evaluate edge consistency around the foreground subject.\\[1pt]
\textit{Criteria:}
\begin{itemize}[nosep,leftmargin=12pt,label=\textbullet]
\item Halo/fringing (bright outline), color spill (green/blue edges), jagged edges
\item Temporal edge stability: edges should not crawl/shimmer frame-to-frame
\item Fine structures: hair, fingers, thin objects, semi-transparent regions
\item Holes/missing parts: accidental cut-outs, broken silhouettes
\item Motion blur handling: edges during fast motion should look natural
\end{itemize}
\textit{Scoring:}
10: Clean \& stable $|$
8--9: Minor noise $|$
6--7: Noticeable $|$
4--5: Strong issues $|$
1--3: Unusable
\end{tcolorbox}

\begin{tcolorbox}[metricbox]
\textbf{Foreground Preservation} (\texttt{FG\_Preserve}) \hfill \textit{Role: Expert evaluator for FG preservation quality.}\\[2pt]
\textit{Input:} (1)~Reference video: original foreground on green screen; (2)~Generated video: foreground extracted from background-replaced composite via mask, also on green screen.\\[2pt]
\textit{Task:} Judge how faithfully the generated video preserves the original foreground. Color/lighting shifts for background matching are \textbf{acceptable} and should \textbf{not} be penalized. Minor compression artifacts or slight edge differences along the green-screen boundary should also be ignored.\\[2pt]
\textit{Criteria:}
\begin{itemize}[nosep,leftmargin=12pt,label=\textbullet]
\item Overall identity: clearly the same person(s)/object(s)
\item Shape \& structure: body proportions, limbs, silhouettes, clothing preserved
\item Motion \& pose: poses and movements match across both videos
\item Texture \& detail: face, clothing patterns, accessories preserved
\item Temporal consistency: no excessive flicker or jitter in the generated foreground
\end{itemize}
\textit{Scoring:}\\
10: Virtually identical foreground---same identity, shape, motion, details; only color/lighting differs.\\
8--9: Very faithful---same person(s), same motion, minor local differences in fine details.\\
6--7: Good preservation---clearly same subject(s) and motion, some noticeable texture loss or slight shape changes.\\
4--5: Moderate changes---recognizable but with visible structural deformations or altered details.\\
2--3: Major changes---significant alterations: missing/added body parts, wrong poses, severe distortions.\\
1: Completely different---the foreground is unrecognizable or entirely replaced.
\end{tcolorbox}

\end{tcolorbox}
\caption{\textbf{Gemini evaluation prompt templates (Cine-NBG Subset, part 2/2).} Edge\_Cons evaluates matte/edge quality of the generated video. FG\_Preserve compares the foreground region in the generated composite with the original foreground reference extracted from the green-screen video.}
\label{fig:prompt_wogt_fg}
\end{figure}

\section{Details on Motion Fidelity Evaluation}
\label{sec:suppl_cotracker}

To rigorously evaluate the temporal motion consistency between the generated videos and the ground truth (GT) in the Cine-Restore subset, we employ CoTracker3~\cite{cotracker3} to establish dense, long-term point correspondences.

\textbf{Point Sampling and Tracking.}
Instead of relying on sparse keypoint detectors, which may fail in unstructured regions, we initialize a uniform $20\times20$ point grid on the first frame of both videos. To ensure exact spatial alignment before tracking, both the generated and GT videos are spatially resized to a consistent resolution (short side scaled to 384 pixels) and temporally synchronized. We track these points across all frames to obtain trajectory coordinates and visibility confidences. To prevent occlusion or out-of-bounds points from skewing the evaluation, we apply a strict \textit{joint visibility mask}. Specifically, a point's trajectory is only considered valid for comparison if its visibility confidence remains greater than 0.5 in \textit{both} the generated and GT videos simultaneously.

\begin{figure*}[!t]
    \centering
    \animategraphics[width=\linewidth]{2}{compare_}{00}{09}
    \caption{\textbf{Visual results of excluded open-source IV2V baselines.} When tasked with green-screen background replacement given a reference background (Column 1) and an input foreground video (Column 2), existing models completely fail to synthesize a coherent video. They suffer from severe hallucination, identity loss, semantic misinterpretation of the green screen, or complete conditioning collapse into noise. Readers can click and play the video clips in this figure using {\color{red}\textbf{Adobe Acrobat}}.}
    \label{fig:suppl_bad_baselines}
\end{figure*}

\textbf{Scale-Invariant Trajectory Comparison.}
A key challenge raised when evaluating motion fidelity across diverse cinematic scenes is the massive variation in object depth, camera focal length, and absolute motion magnitude. To ensure a fair and normalized comparison across completely different scenes, we evaluate the relative displacement trajectories (i.e., coordinate offsets relative to the first frame) using \textbf{Trajectory Cosine Similarity}. Formally, let $P_{t}^{(i)}\in\mathbb{R}^2$ and $\hat{P}_{t}^{(i)}\in\mathbb{R}^2$ denote the 2D spatial coordinates of the $i$-th tracked point at frame $t$ for the GT and generated videos, respectively. The temporal displacement relative to the first frame is defined as $\Delta P_{t}^{(i)}=P_{t}^{(i)}-P_{0}^{(i)}$ (and analogously for $\Delta\hat{P}_{t}^{(i)}$). Let $V^{(i)}$ be the set of valid frames for point $i$, defined by the joint visibility mask. As formulated in Eq.~\ref{eq:cos_sim}, the trajectory cosine similarity for the $i$-th point is computed by flattening the spatial displacements over time:
\begin{equation}
S^{(i)}=\frac{\sum_{t\in V^{(i)}}\langle\Delta P_{t}^{(i)},\Delta\hat{P}_{t}^{(i)}\rangle}{\sqrt{\sum_{t\in V^{(i)}}\|\Delta P_{t}^{(i)}\|_2^2}\sqrt{\sum_{t\in V^{(i)}}\|\Delta\hat{P}_{t}^{(i)}\|_2^2}},
\label{eq:cos_sim}
\end{equation}
where $\langle\cdot,\cdot\rangle$ denotes the inner product. As shown in Eq.~\ref{eq:motion_score}, the final motion consistency score is obtained by averaging the similarity across all $N_{valid}$ valid points:
\begin{equation}
\text{Motion Cons.}=\frac{1}{N_{valid}}\sum_{i=1}^{N_{valid}}S^{(i)}.
\label{eq:motion_score}
\end{equation}
This metric strictly evaluates the directional and structural fidelity of the motion path, making it naturally invariant to the absolute pixel distance traveled.

\section{Correlation between Human Preference and Automated Evaluation}
\label{sec:suppl_correlation}

To rigorously validate the reliability of our Gemini-based automated evaluation protocol, we compute the statistical correlation between the automated scores and the human user study votes.

\textbf{Formulation.} In our user study, participants evaluated the methods in a 6-way choice setup. For a specific evaluation metric, let $u_{v,m}$ denote the number of human votes method $m$ received for video $v$. We convert these votes into a comparative human preference distribution:
\begin{equation}
    p_{v,m} = \frac{u_{v,m}}{\sum_{j=1}^{6} u_{v,j}}.
\end{equation}
Similarly, let $g_{v,m} \in [1, 10]$ denote the absolute score assigned by Gemini for the same method and video. To map these absolute scores into a relative preference distribution that intrinsically aligns with the 6-way user study format, we apply a softmax function:
\begin{equation}
    q_{v,m} = \frac{\exp(g_{v,m})}{\sum_{j=1}^{6} \exp(g_{v,j})}.
\end{equation}
For each metric, we flatten these distributions across all evaluated videos and the 6 methods into comprehensive vectors $U$ and $G$ of length $N$. Let $R(U)$ and $R(G)$ denote the assigned fractional ranks of the elements in $U$ and $G$, respectively. The Spearman rank correlation coefficient ($\rho$) is then computed as the Pearson correlation between these rank variables:
\begin{equation}
    \rho = \frac{\sum_{i=1}^{N} (R(U_i) - \bar{R}_U)(R(G_i) - \bar{R}_G)}{\sqrt{\sum_{i=1}^{N} (R(U_i) - \bar{R}_U)^2 \sum_{i=1}^{N} (R(G_i) - \bar{R}_G)^2}},
\end{equation}
where $\bar{R}_U$ and $\bar{R}_G$ are the mean ranks.

\begin{figure}[!t]
    \centering
    \includegraphics[width=\linewidth]{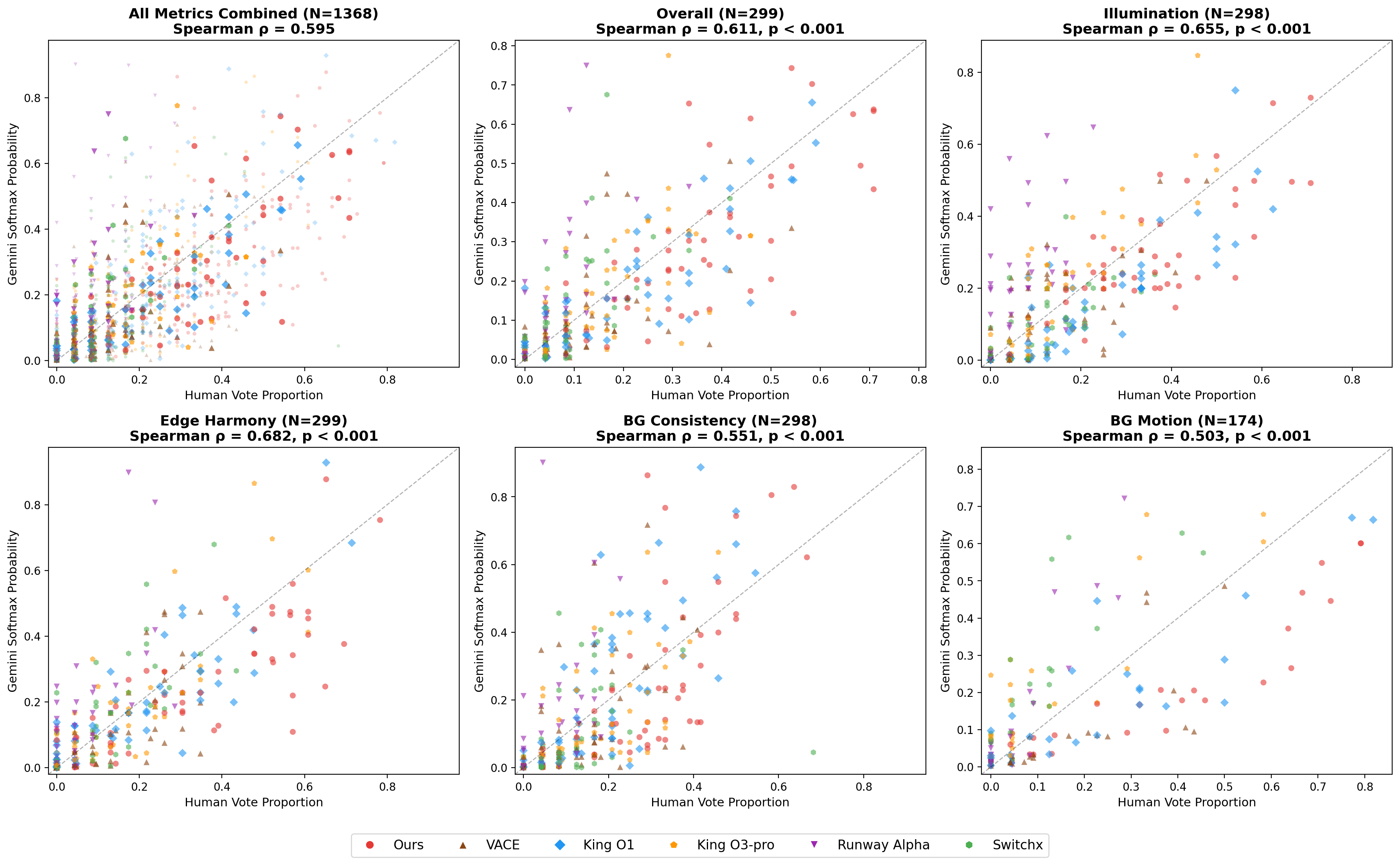}
    \caption{\textbf{Correlation between human preference and Gemini evaluation.} The scatter plots illustrate the relationship between the human vote proportion and the Gemini softmax probability for each method. The Spearman rank correlation ($\rho$) ranges from 0.503 to 0.682 across all individual dimensions, with an overall combined correlation of 0.595 (all with $p < 0.001$). This indicates a robust and statistically significant alignment between the automated judge and human consensus on this highly subjective task.}
    \label{fig:suppl_correlation}
\end{figure}

\textbf{Results and Analysis.}
As illustrated in Fig.~\ref{fig:suppl_correlation}, we compute the correlation across all key dimensions. It is important to note that for the \textit{Background Motion Rationality} metric, the correlation is computed over 29 videos ($N = 29 \times 6 = 174$ instances). This is because the remaining 21 videos in our user study belong to the Cine-Restore subset, which possess ground-truth videos and are strictly evaluated using the dense point-tracking metric (CoTracker3, as detailed in Sec.~\ref{sec:suppl_cotracker}) for motion fidelity, rather than using Gemini. For all other semantic metrics, the correlation is computed across the full set of videos ($N \approx 300$).

Assessing the perceptual quality of high-resolution cinematic video generation is an inherently subjective task, heavily influenced by individual aesthetic preferences, varying sensitivities to specific artifacts, and complex foreground-background dynamics. Given this high degree of subjectivity and the inherently noisy nature of human voting in a 6-way choice setup, achieving a Spearman correlation coefficient ($\rho$) ranging from 0.503 to 0.682 (with strict statistical significance $p < 0.001$) is considered a strongly robust alignment. These results validate that our Gemini-based automated pipeline reliably reflects aggregate human consensus and serves as a trustworthy proxy for large-scale perceptual quality assessment.

\begin{figure*}[t!]
    \centering
    \includegraphics[width=\linewidth]{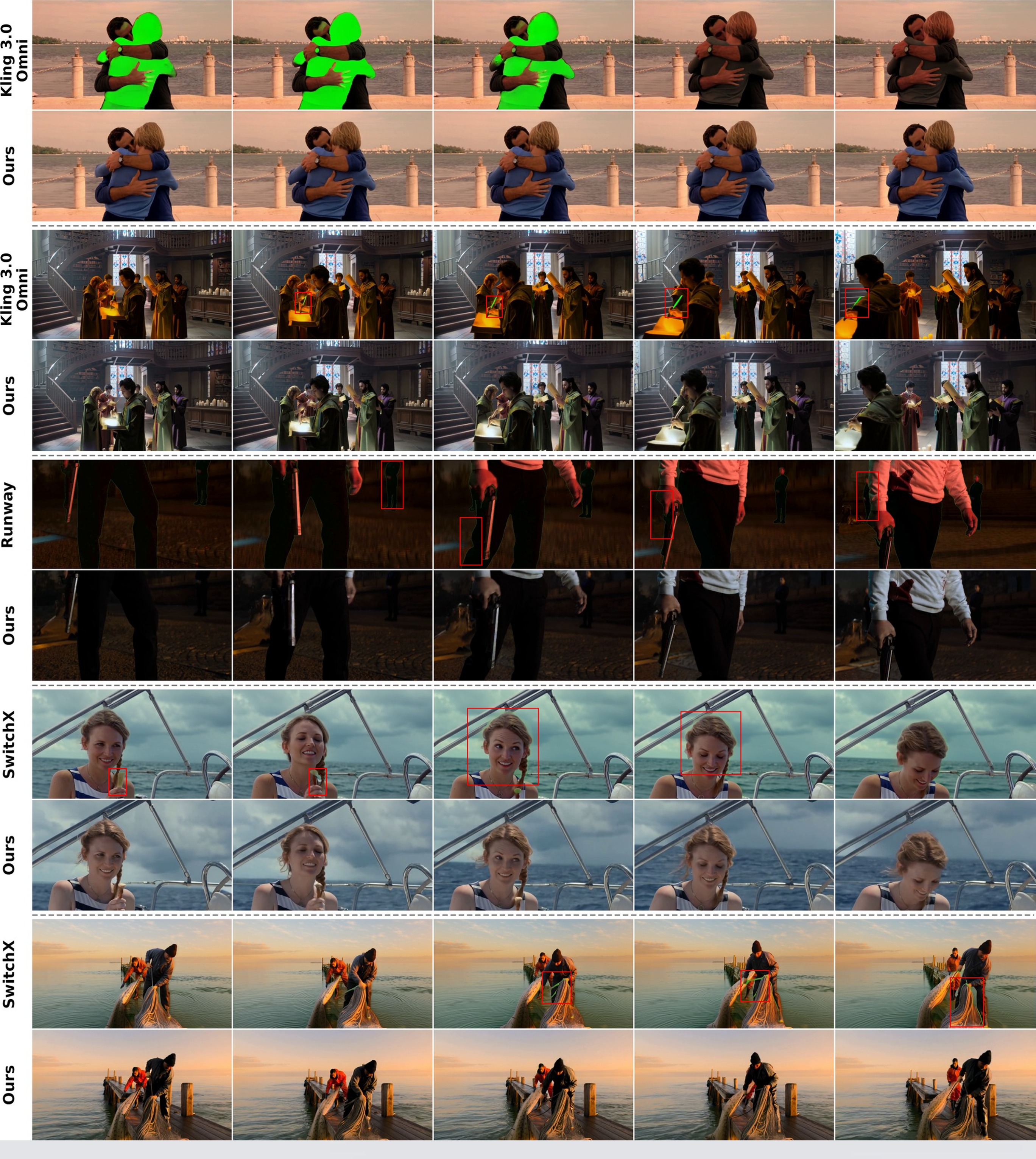}
    \caption{\textbf{Superior edge harmonization.} Compared to baselines that suffer from severe green spill and boundary artifacts (highlighted by red boxes), our model generates clean, natural edges without residual green contamination.}
    \label{fig:suppl_edge}
\end{figure*}

\begin{figure*}[t!]
    \centering
    \includegraphics[width=\linewidth]{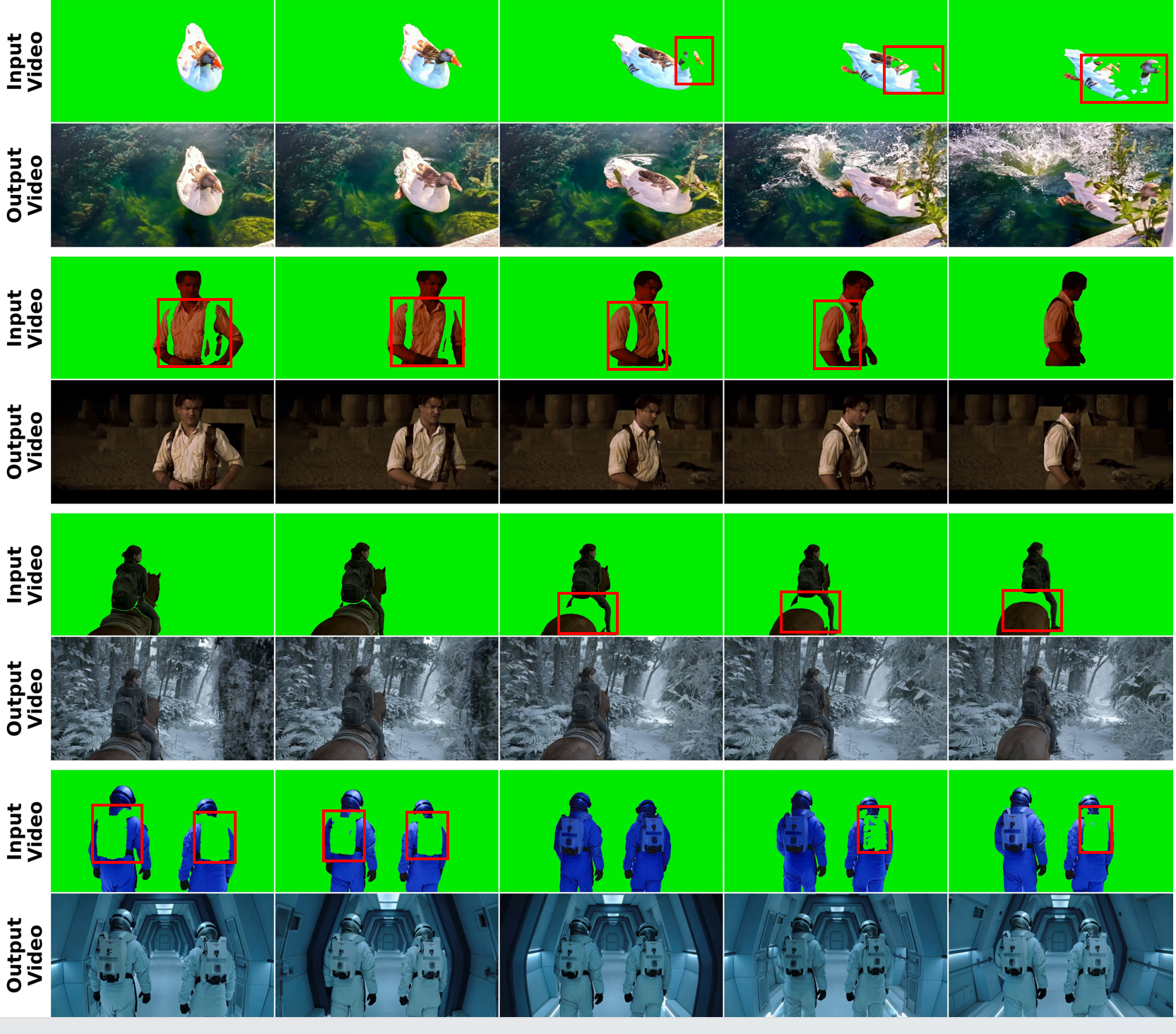}
    \caption{\textbf{Robustness to imperfect foreground segmentation.} We highlight four challenging cases where the input green-screen videos (top rows) contain mask defects, \textbf{as indicated by the red bounding boxes}. These include severely corrupted body parts, arbitrary holes, and artificial occlusions. Without any explicit inpainting prompts, our model (bottom rows) robustly reconstructs the missing foreground structures while seamlessly compositing them into the generated backgrounds with global illumination and structural harmony.}
    \label{fig:suppl_robustness}
\end{figure*}

\begin{figure*}[!t]
    \centering
    \includegraphics[width=0.9\linewidth]{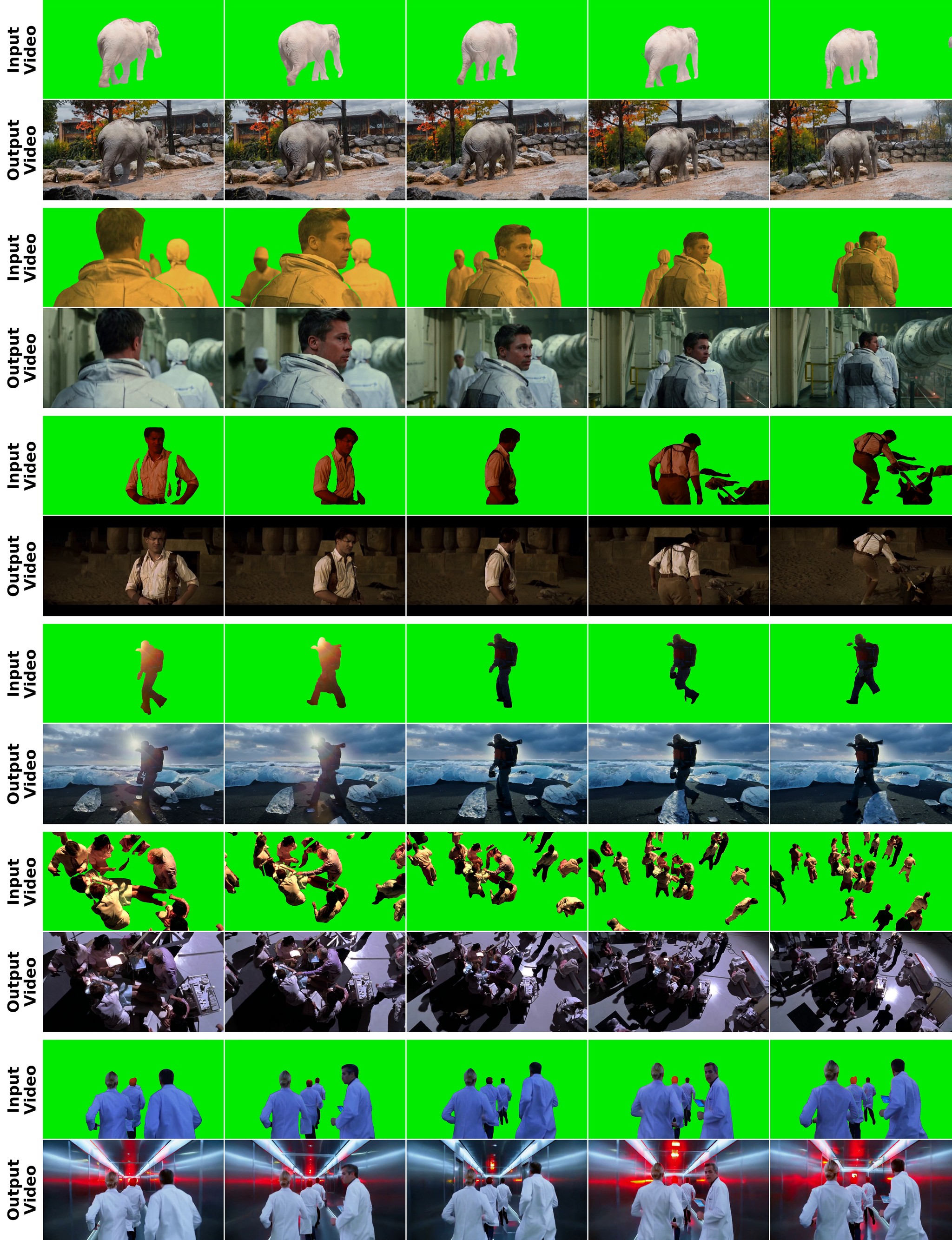}
    \caption{\textbf{Implicit scene-adaptive relighting.} Our model automatically harmonizes the foreground illumination with the newly synthesized backgrounds without requiring any explicit relighting prompts or maps.}
    \label{fig:suppl_relighting}
\end{figure*}

\begin{figure*}[t!]
    \centering
    \includegraphics[width=\linewidth]{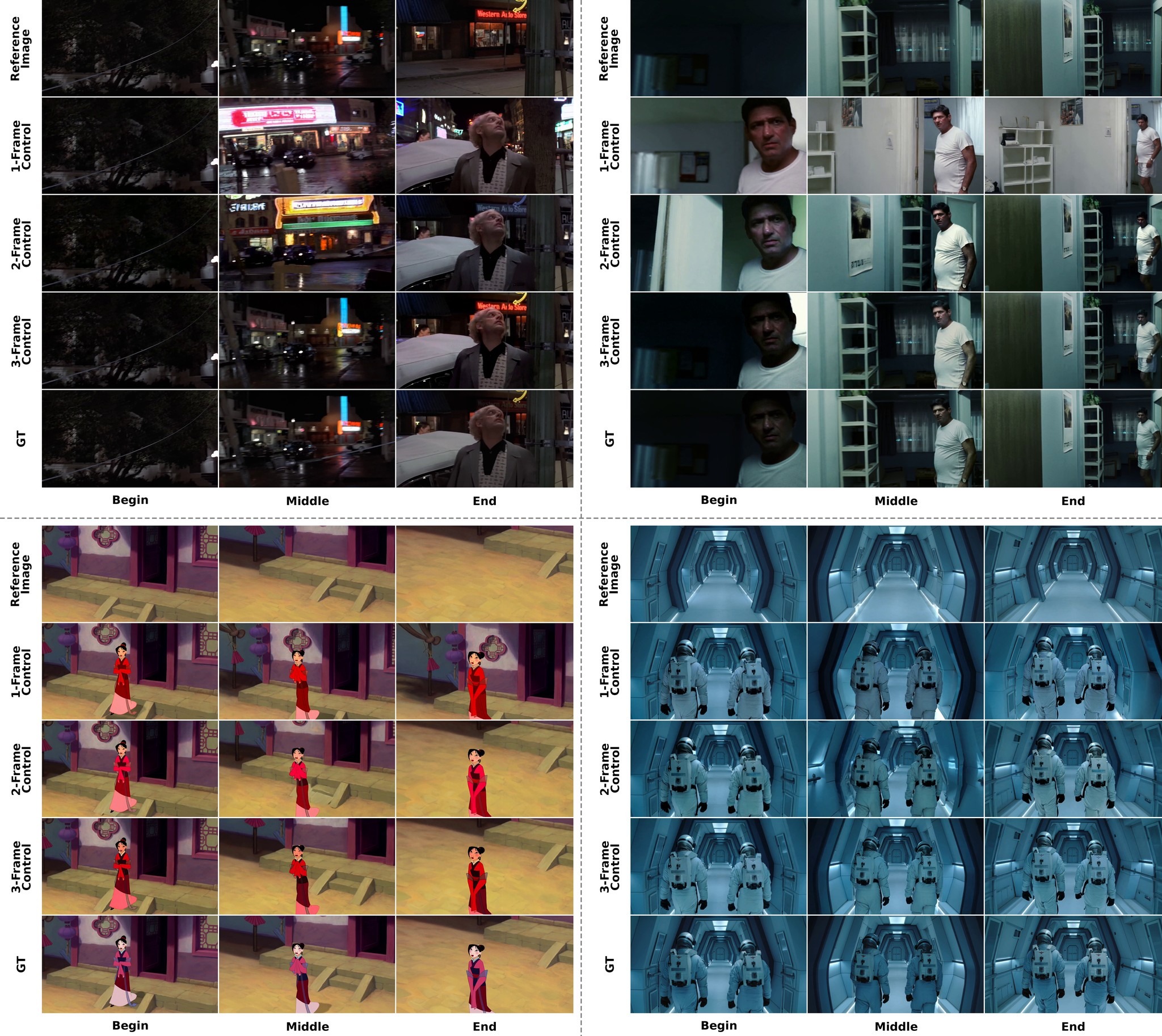}
    \caption{\textbf{Effect of multi-frame background control.} Comparing inference results conditioned on 1, 2, and 3 background reference images. The 3-reference control effectively anchors the background at the beginning, middle, and end temporal locations, yielding the most temporally coherent results that closely match the ground truth (GT). Fewer reference frames lead to information deficits at unconditioned time steps, causing structural deviations and abrupt disappearance or morphing of background objects.}
    \label{fig:suppl_multiframe}
\end{figure*}

\begin{figure*}[t!]
    \centering
    \includegraphics[width=\linewidth]{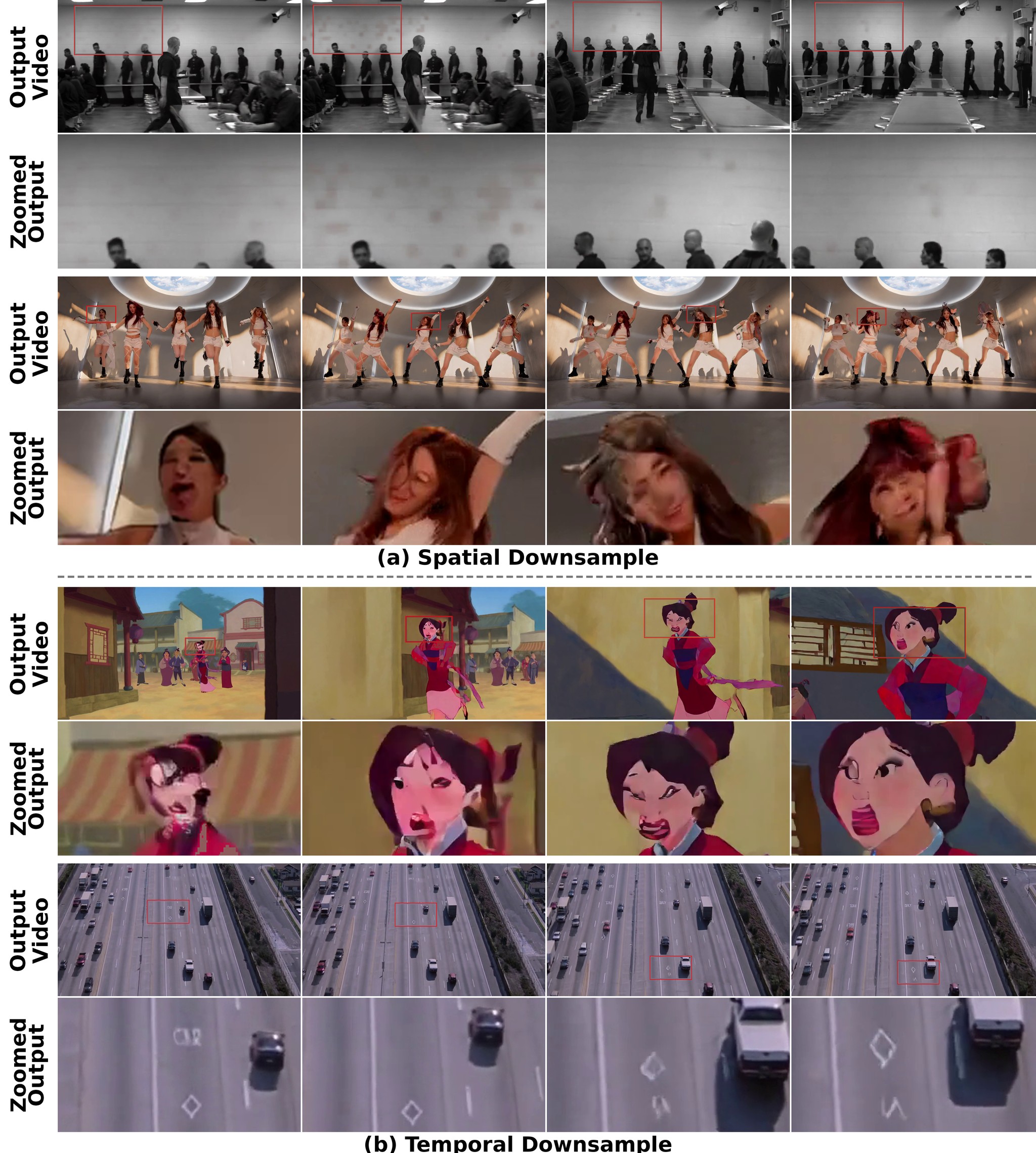}
    \caption{\textbf{Visual demonstration of the limitations of efficient conditioning strategies.} Both spatial downsampling (a) and temporal downsampling (b) introduce noticeable visual degradation. Key artifacts include severe facial degradation, the emergence of black blotches in the background, and the deformation or disappearance of structural symbols.}
    \label{fig:suppl_downsample}
\end{figure*}

\section{Analysis of Excluded Open-Source Baselines}
\label{sec:suppl_bad_baselines}

To ensure a fair and rigorous comparison under the exact same conditioning setting, we first strictly exclude text-to-video (T2V) or purely image-to-video (I2V) models that lack Image-Video-to-Video (IV2V) support. Specifically, these models cannot perform video background replacement conditioned on one or more background reference images. Since their operational setting fundamentally differs from ours, treating them as direct baselines would be intrinsically unfair.

Following this criterion, we extensively surveyed and tested existing open-source models capable of IV2V generation or video editing, including AnyV2V~\cite{anyv2v}, LoRAEdit~\cite{loraedit}, MagicAnimate~\cite{magicanimate}, runway-stable-diffusion-inpainting (SD-Inpaint)~\cite{sdinpaint}, VideoComposer~\cite{videocomposer}, and VideoPainter~\cite{videopainter}. However, empirical testing reveals that these general-purpose editing models perform exceptionally poorly on the demanding task of green-screen cinematic background replacement.

As visually demonstrated in Fig.~\ref{fig:suppl_bad_baselines}, these models fail to generate temporally coherent or structurally normal videos, exhibiting several critical failure modes:

\begin{itemize}
    \item \textbf{Conditioning Collapse (e.g., VideoComposer, VideoPainter):} These models completely fail to fuse the high-resolution foreground and background conditions, often collapsing into severe color shifts, purple/blue artifacts, or pure chaotic noise, losing all structural information of the input.
    \item \textbf{Foreground Identity Loss (e.g., MagicAnimate, AnyV2V, LoRAEdit):} Instead of preserving the driving foreground, these models suffer from severe temporal warping and hallucinate entirely incorrect subjects or geometries, completely destroying the identity and motion of the original actor.
    \item \textbf{Semantic Misinterpretation (e.g., SD-Inpaint):} Mask-based or general inpainting approaches fail to physically understand the green-screen compositing task. They often treat the green backdrop as a literal semantic object to be retained or paint erratic artifacts around the foreground boundaries, resulting in severe flickering and green spill.
\end{itemize}

In conclusion, because none of these surveyed IV2V models can successfully generate a basic, visually intelligible composite video under our heterogeneous conditioning setup, including them in the quantitative benchmarks or user studies would be uninformative. Therefore, they are excluded from our main evaluation as unsuitable direct baselines.

\section{Superior Edge Harmonization and Green Spill Suppression}
\label{sec:suppl_edge}

A fundamental challenge in cinematic background replacement is managing the boundaries of the foreground subjects. Baseline video editing models frequently struggle to fully eliminate the original green screen, leading to a phenomenon known as ``green spill'' or boundary color contamination.

As visually demonstrated in Fig.~\ref{fig:suppl_edge}, existing commercial APIs---\emph{occasionally} Kling 3.0 Omni and \emph{more consistently} Runway and Switchx---fail to synthesize clean edges. They either leave noticeable green halos around fine details (e.g., hair, hands, and thin objects), or, in more severe failure cases, hallucinate massive green color blocks that completely obscure the foreground subjects (\textbf{highlighted by red boxes}).

In contrast, our generative framework intrinsically understands the semantic boundaries and robustly suppresses any residual green screen artifacts. It synthesizes clean, precise, and naturally blended edges, ensuring that the foreground seamlessly integrates into the new environments without any structural distortion or color contamination.

\section{Robustness to Imperfect Foreground Segmentation}
\label{sec:suppl_robustness}

In real-world cinematic production and automated video matting, foreground segmentation masks are frequently imperfect, often suffering from local holes, missing limbs, or artificial occlusions (e.g., tracking markers or rigging equipment). Traditional compositing pipelines and naive editing models are highly sensitive to these defects, as they directly transfer the missing regions or mask errors into the final composite.

In contrast, our proposed model exhibits remarkable zero-shot robustness against segmentation artifacts. Because our framework operates within a generative latent space and leverages the strong structural priors of the diffusion backbone, it does not merely copy-and-paste the foreground. Instead, it performs joint background replacement and semantic foreground inpainting, achieving \textbf{global visual and temporal coherence}. By comprehensively capturing the global context of the scene, the model ensures that the locally reconstructed foreground details are structurally, temporally, and photometrically consistent with the overall environmental lighting and geometry, rather than appearing as isolated patches.

As visually demonstrated in Fig.~\ref{fig:suppl_robustness}, we deliberately test our model on input foreground videos with some degradation. Even when confronted with prominent masked-out regions---such as the missing head of the swimming duck, the severely corrupted torso of the human actor, truncated animal bodies, or large artificial holes on the astronauts' space suits (\textbf{highlighted by red boxes})---our model accurately deduces the underlying semantic geometry and texture. It seamlessly hallucinates the missing foreground elements, maintains temporal consistency across frames, and simultaneously harmonizes the illumination with the newly synthesized background. This robust self-correction capability significantly relaxes the strict requirements for flawless upstream matting, making our pipeline highly practical for real-world deployment.

\section{Implicit Visual Relighting and Scene Adaptation}
\label{sec:suppl_relighting}

As illustrated in Fig.~\ref{fig:suppl_relighting}, our method implicitly relights foreground subjects to harmonize with newly synthesized environments, achieving scene-adaptive illumination without any explicit relighting prompts or maps. Instead of merely pasting the foreground, the model accurately generates context-aware shadows, color hues, and complex rim-lighting that strictly match the lighting conditions of the new backgrounds. This zero-shot relighting capability ensures the photorealistic integration essential for high-end cinematic compositing.

\tcbset{
  fieldbox/.style={
    colback=gray!5, colframe=gray!50,
    boxrule=0.4pt, arc=1.5pt, left=3pt, right=3pt, top=2pt, bottom=2pt,
    before skip=3pt, after skip=3pt,
  }
}

\section{Gemini-based Structured Captioning Details}
\label{sec:suppl_gemini_caption}

\begin{figure}[!t]
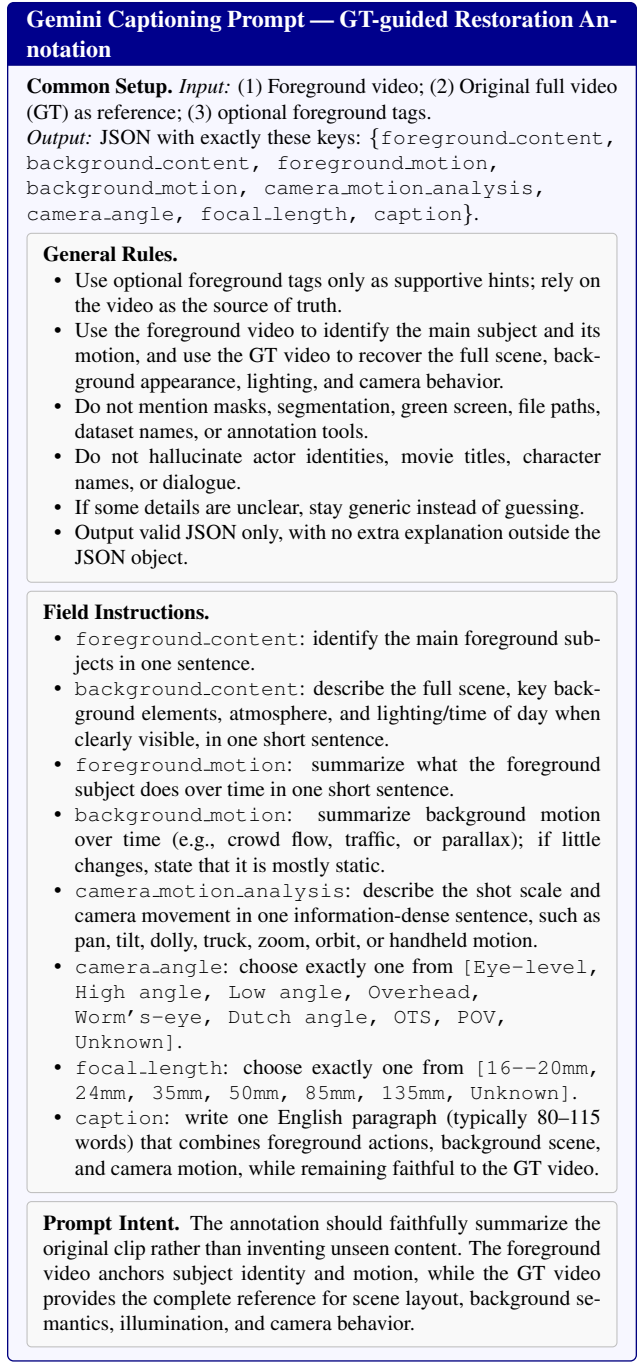

\begin{tcolorbox}[promptbox, title=Gemini Captioning Prompt --- GT-guided Restoration Annotation]
\footnotesize

\textbf{Common Setup.}
\textit{Input:} (1)~Foreground video; (2)~Original full video (GT) as reference; (3)~optional foreground tags.\\
\textit{Output:} JSON with exactly these keys:
\texttt{\{foreground\_content, background\_content, foreground\_motion, background\_motion, camera\_motion\_analysis, camera\_angle, focal\_length, caption\}}.

\begin{tcolorbox}[fieldbox]
\textbf{General Rules.}
\begin{itemize}[nosep,leftmargin=12pt,label=\textbullet]
\item Use optional foreground tags only as supportive hints; rely on the video as the source of truth.
\item Use the foreground video to identify the main subject and its motion, and use the GT video to recover the full scene, background appearance, lighting, and camera behavior.
\item Do not mention masks, segmentation, green screen, file paths, dataset names, or annotation tools.
\item Do not hallucinate actor identities, movie titles, character names, or dialogue.
\item If some details are unclear, stay generic instead of guessing.
\item Output valid JSON only, with no extra explanation outside the JSON object.
\end{itemize}
\end{tcolorbox}

\begin{tcolorbox}[fieldbox]
\textbf{Field Instructions.}
\begin{itemize}[nosep,leftmargin=12pt,label=\textbullet]
\item \texttt{foreground\_content}: identify the main foreground subjects in one sentence.
\item \texttt{background\_content}: describe the full scene, key background elements, atmosphere, and lighting/time of day when clearly visible, in one short sentence.
\item \texttt{foreground\_motion}: summarize what the foreground subject does over time in one short sentence.
\item \texttt{background\_motion}: summarize background motion over time (e.g., crowd flow, traffic, or parallax); if little changes, state that it is mostly static.
\item \texttt{camera\_motion\_analysis}: describe the shot scale and camera movement in one information-dense sentence, such as pan, tilt, dolly, truck, zoom, orbit, or handheld motion.
\item \texttt{camera\_angle}: choose exactly one from \texttt{[Eye-level, High angle, Low angle, Overhead, Worm's-eye, Dutch angle, OTS, POV, Unknown]}.
\item \texttt{focal\_length}: choose exactly one from \texttt{[16--20mm, 24mm, 35mm, 50mm, 85mm, 135mm, Unknown]}.
\item \texttt{caption}: write one English paragraph (typically 80--115 words) that combines foreground actions, background scene, and camera motion, while remaining faithful to the GT video.
\end{itemize}
\end{tcolorbox}

\begin{tcolorbox}[fieldbox]
\textbf{Prompt Intent.}
The annotation should faithfully summarize the original clip rather than inventing unseen content. The foreground video anchors subject identity and motion, while the GT video provides the complete reference for scene layout, background semantics, illumination, and camera behavior.
\end{tcolorbox}

\end{tcolorbox}
\caption{\textbf{Prompt template for GT-guided structured caption generation.} Gemini receives a foreground video together with the original ground truth video as reference, and outputs a structured JSON annotation for the original clip.}
\label{fig:prompt_caption_gt}
\end{figure}

\begin{figure}[!t]
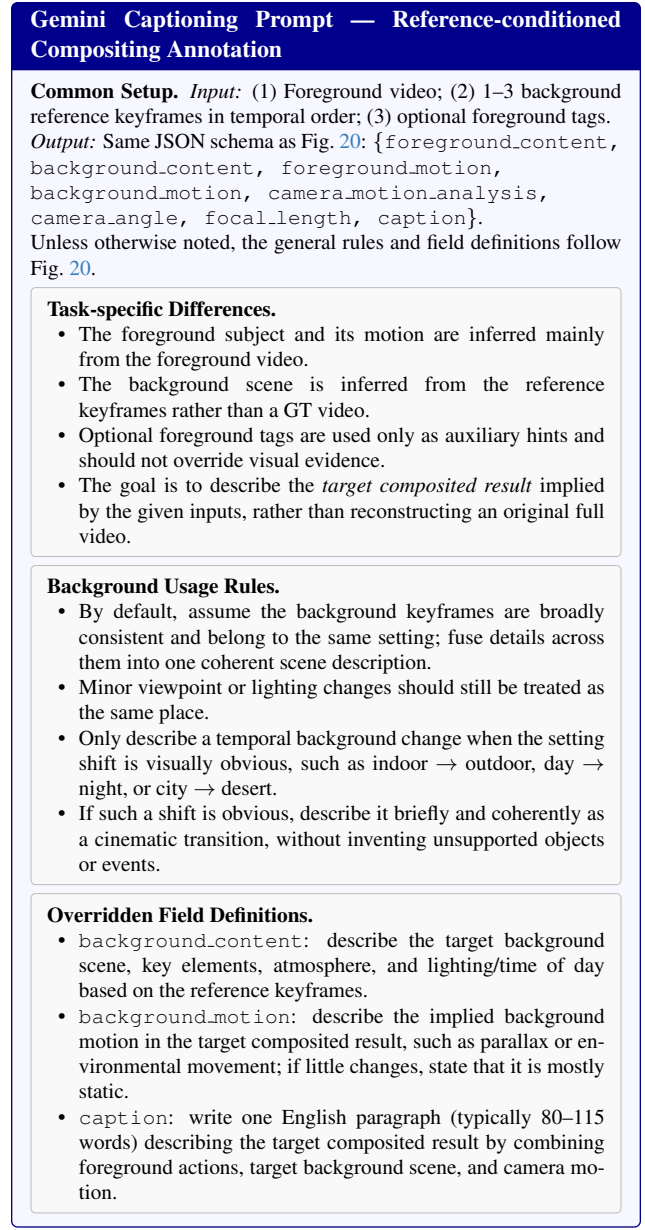

\begin{tcolorbox}[promptbox, title=Gemini Captioning Prompt --- Reference-conditioned Compositing Annotation]
\footnotesize

\textbf{Common Setup.}
\textit{Input:} (1)~Foreground video; (2)~1--3 background reference keyframes in temporal order; (3)~optional foreground tags.\\
\textit{Output:} Same JSON schema as Fig.~\ref{fig:prompt_caption_gt}:
\texttt{\{foreground\_content, background\_content, foreground\_motion, background\_motion, camera\_motion\_analysis, camera\_angle, focal\_length, caption\}}.\\
Unless otherwise noted, the general rules and field definitions follow Fig.~\ref{fig:prompt_caption_gt}.

\begin{tcolorbox}[fieldbox]
\textbf{Task-specific Differences.}
\begin{itemize}[nosep,leftmargin=12pt,label=\textbullet]
\item The foreground subject and its motion are inferred mainly from the foreground video.
\item The background scene is inferred from the reference keyframes rather than a GT video.
\item Optional foreground tags are used only as auxiliary hints and should not override visual evidence.
\item The goal is to describe the \emph{target composited result} implied by the given inputs, rather than reconstructing an original full video.
\end{itemize}
\end{tcolorbox}

\begin{tcolorbox}[fieldbox]
\textbf{Background Usage Rules.}
\begin{itemize}[nosep,leftmargin=12pt,label=\textbullet]
\item By default, assume the background keyframes are broadly consistent and belong to the same setting; fuse details across them into one coherent scene description.
\item Minor viewpoint or lighting changes should still be treated as the same place.
\item Only describe a temporal background change when the setting shift is visually obvious, such as indoor $\rightarrow$ outdoor, day $\rightarrow$ night, or city $\rightarrow$ desert.
\item If such a shift is obvious, describe it briefly and coherently as a cinematic transition, without inventing unsupported objects or events.
\end{itemize}
\end{tcolorbox}

\begin{tcolorbox}[fieldbox]
\textbf{Overridden Field Definitions.}
\begin{itemize}[nosep,leftmargin=12pt,label=\textbullet]
\item \texttt{background\_content}: describe the target background scene, key elements, atmosphere, and lighting/time of day based on the reference keyframes.
\item \texttt{background\_motion}: describe the implied background motion in the target composited result, such as parallax or environmental movement; if little changes, state that it is mostly static.
\item \texttt{caption}: write one English paragraph (typically 80--115 words) describing the target composited result by combining foreground actions, target background scene, and camera motion.
\end{itemize}
\end{tcolorbox}

\end{tcolorbox}
\caption{\textbf{Prompt template for reference-conditioned structured caption generation without GT.} Compared with Fig.~\ref{fig:prompt_caption_gt}, this version replaces the GT video with 1--3 background reference images and asks Gemini to describe the target composited result implied by the inputs.}
\label{fig:prompt_caption_nogt}
\end{figure}

We use Gemini-3-Flash~\cite{gemini} to generate structured video annotations for dataset curation and training. Two complementary protocols are adopted.
(1)~\textbf{GT-guided restoration annotation}: the model receives a foreground video together with the original full video (ground truth), and produces a structured description faithful to the original clip.
(2) \textbf{Reference-conditioned compositing annotation}: in the absence of a ground-truth video, the model receives only a foreground video together with 1--3 background reference images, and must infer a structured caption for the anticipated composited result.
In both protocols, Gemini outputs a JSON record containing foreground/background content, foreground/background motion, camera motion analysis, camera angle, focal length, and a summary caption. The full prompt templates are shown in Fig.~\ref{fig:prompt_caption_gt} and Fig.~\ref{fig:prompt_caption_nogt}.

\section{Effect of Multi-Frame Control}
\label{sec:suppl_multiframe}

To further investigate the impact of multi-frame background conditioning, we compare the inference results using 1, 2, and 3 background reference images as control signals. As visually demonstrated in Fig.~\ref{fig:suppl_multiframe}, the 3-frame control achieves the best synthesis quality. By accurately injecting spatial-temporal background information at the beginning, middle, and end locations via temporal positional encoding cloning, the model maintains the highest temporal consistency and coherence, closely matching the ground truth (GT). Conversely, reducing the number of background reference images leads to a deficit of background information at the unconditioned temporal positions. As a result, the generated videos exhibit noticeable deviations from the GT.

Without explicit guidance at these crucial temporal anchors, the model struggles to reconstruct unseen regions under large camera motions, resulting in bizarre background variations---such as background objects suddenly disappearing, morphing, or the model hallucinating entirely inconsistent content.

\begin{figure*}[!t]
  \centering
  \animategraphics[width=\linewidth]{5}{fail_case_output_v6/fail_case_}{00}{07}
  \caption{\textbf{Failure case examples.} We highlight three typical failure modes: (1) artifacts near transparent/refractive boundaries (e.g., glass edges), (2) occasional abnormal foreground relighting (stochastic; alleviated by re-sampling seeds), and (3) unrealistic background details under abrupt scene transitions with weak semantic continuity. Readers can click and play the video clips in this figure using {\color{red}\textbf{Adobe Acrobat}}.}
  \label{fig:fail_cases}
\end{figure*}

\section{Explorations on Efficient Conditioning}
\label{sec:efficient}

Directly conditioning on a high-resolution foreground video $V^{fg}$ across all frames (e.g., 1080p, 121 frames) is computationally expensive. Before arriving at our final design, we empirically explored two straightforward efficiency strategies to reduce the input condition redundancy:

\textbf{(1) Temporal Subsampling (Joint Interpolation):} We temporally subsampled the input foreground video $V^{fg}$ with a stride of 2, reducing the condition sequence to 61 frames while maintaining the 1080p resolution. The model was trained to output the full 121-frame video $V^{out}$ at 1080p, essentially forcing the network to perform joint Video Frame Interpolation (VFI) and background replacement.

\textbf{(2) Spatial Downsampling (Joint Super-Resolution):} We spatially downsampled the input foreground video $V^{fg}$ to 720p while preserving the full temporal density (121 frames). The target output $V^{out}$ remained at 1080p, implicitly requiring the model to perform joint Video Super-Resolution (VSR) alongside the core compositing task.

\textbf{Analysis of Limitations.} Empirically, both naive strategies yielded suboptimal results, as visually demonstrated in Fig.~\ref{fig:suppl_downsample}. Specifically, we observed three critical failure modes: (1) structural distortions and texture degradation in fine-grained foreground regions, most notably resulting in severe \textbf{facial degradation}; (2) the emergence of noticeable \textbf{black spots and blotchy artifacts} in the synthesized background areas; and (3) the \textbf{disappearance or deformation of symbols} and structural markings (e.g., road signs and ground texts). We hypothesize that this degradation stems from a critical information bottleneck: forcing the network to simultaneously hallucinate missing spatial or temporal details (via VFI or VSR) while attempting complex background synthesis and scene-adaptive relighting overloads the model's capacity. The uniform reduction of spatial or temporal resolution inevitably discards critical identity cues that cannot be perfectly recovered.

\textbf{Final Strategy: Full-Resolution Symmetric Processing.}
Due to the degradation in facial details and foreground fidelity, we discarded these asymmetrical input-output compression schemes. In our final pipeline, both the foreground condition (encoded as $Z^{fg}$ from $V^{fg}$) and the denoising video (represented by $Z_s$ and decoded to $V^{out}$) strictly maintain the exact same high resolution and frame count (i.e., 1080p, 97 frames). Although this symmetrical full-resolution processing increases the computational and memory footprint (as detailed in Sec.~\ref{sec:suppl_cost}), we conclude that it is strictly necessary to preserve identity and achieve cinematic-level synthesis quality.

\section{Computational Cost and Resources}

\label{sec:suppl_cost}

\textbf{Training Details and Cost.} Our model was trained directly on high-resolution cinematic sequences (1080p resolution, 97 frames) using 8 NVIDIA H200 GPUs for approximately 5 days, reaching a total of 6,000 training steps. During this phase, the peak VRAM consumption was around 135~GB per GPU. To enhance the model's robustness and condition flexibility, we employ a stochastic frame dropping strategy during training: each background condition frame is independently retained with a probability of 50\%. Consequently, the network learns a robust generative prior capable of performing high-quality video synthesis conditioned on a variable number of reference frames (typically ranging from 0 to 3).

\noindent
\textbf{Inference Cost.} To evaluate the practical deployment of our method, we report the inference latency and memory footprint. Unlike the stochastic selection used in training, inference is performed by deterministically specifying a fixed number of condition frames to strictly guide the generation. Running our WAN~2.2-5B~\cite{wan2025} based model on a single NVIDIA H200 GPU to generate a high-resolution cinematic sequence (1080p resolution, 97 frames) with 50 denoising steps requires approximately 25 minutes and consumes around 100~GB of VRAM. While we acknowledge that this inference process is relatively slow and computationally intensive, our primary objective is to achieve cinematic-level visual quality and strict temporal coherence. At the current stage, prioritizing zero-shot high-fidelity visual effects over speed is a deliberate and necessary trade-off. To address the runtime and memory constraints for real-world applications, our immediate future work involves applying progressive distillation techniques to compress the current model into a few-step (e.g., 4-step) generation framework, which will significantly accelerate inference without heavily compromising the synthesis quality.

\section{Failure Cases Analysis}
\label{sec:fail}

Despite strong overall performance, our method still exhibits several failure modes, mainly in scenarios that require (i) view-dependent optical effects, (ii) stable illumination reasoning under large foreground--background discrepancy, or (iii) robust synthesis under abrupt background transitions. Fig.~\ref{fig:fail_cases} summarizes three representative examples.

\textbf{(1) Complex physical effects.}
When the camera moves across the boundary of a glass table, the observed appearance contains strong transmission/refraction distortions that are highly view-dependent and deviate from standard alpha compositing and Lambertian assumptions.
As a result, the model may produce local structural breaks or texture corruption around the glass-edge region, especially when the foreground interacts with thin or transparent boundaries.

\textbf{(2) Occasional abnormal relighting artifacts.}
In a small portion of cases, we observe implausible local highlights or specular blobs on the foreground. This issue is stochastic and can often be mitigated by re-sampling different random seeds and providing multiple candidates to users.
We hypothesize that large illumination discrepancy between foreground and background, together with rare challenging training examples, may cause the model to incorrectly transfer strong lighting patterns to the foreground.

\textbf{(3) Abrupt background changes with weak semantic continuity.}
When the background undergoes drastic scene changes and adjacent frames have little perceptual correspondence, the model needs to hallucinate large unseen regions under heavy disocclusion.
This increases synthesis difficulty and may lead to unrealistic background details and degraded temporal coherence. We believe such extreme transitions are under-represented in training, where most background variations still preserve a traceable continuity.